\documentclass[preprint,12pt]{elsarticle}

\usepackage{amsmath,amsfonts}
\usepackage{algorithmic}
\usepackage{array}
\usepackage[tight,footnotesize]{subfigure}
\usepackage{textcomp}
\usepackage{stfloats}
\usepackage{url}
\usepackage{verbatim}
\usepackage{graphicx}
\def\BibTeX{{\rm B\kern-.05em{\sc i\kern-.025em b}\kern-.08em
    T\kern-.1667em\lower.7ex\hbox{E}\kern-.125emX}}
\usepackage{balance}
\usepackage{booktabs}
\usepackage{caption}
\usepackage{makecell}
\usepackage{adjustbox}
\usepackage{wrapfig}
\usepackage[linesnumbered,ruled]{algorithm2e}
\usepackage{amssymb}
\usepackage{lineno}
\usepackage{float}
\usepackage{algorithmic}
\usepackage[tight,footnotesize]{subfigure}
\usepackage{booktabs}
\usepackage{caption}
\usepackage{makecell}
\usepackage{adjustbox}
\usepackage{wrapfig}
\usepackage{xcolor}
\usepackage[linesnumbered,ruled]{algorithm2e}
\usepackage{listings}
\usepackage{makecell}
\usepackage{pifont}
\newcommand{\cmark}{\ding{51}}%
\newcommand{\xmark}{\ding{55}}%

\journal{Journal Name}

\begin{document}
\sloppy
\setlength{\parskip}{0pt}

\begin{frontmatter}

\title{MindGap: A Conversational AI Framework for Upstream Neuroplastic Intervention in Post-Traumatic Stress Disorder}

\author[label1]{Eranga Bandara}
\ead{cmedawer@odu.edu}

\author[label1]{Ross Gore}
\ead{rgore@odu.edu}

\author[label10]{Asanga Gunaratna}
\ead{asanga.gunaratna@complianceoslab.app}

\author[label1]{Ravi Mukkamala}
\ead{rmukkama@odu.edu}

\author[label15]{Nihal Siriwardanagea}
\ead{nihal@gsi2.com}

\author[label7]{Sachini Rajapakse}
\ead{sachini.rajapakse@iciclelabs.ai}

\author[label7]{Isurunima Kularathna}
\ead{isurunima.kularathna@iciclelabs.ai}

\author[label7]{Pramoda Karunarathna}
\ead{pramoda.karunarathna@iciclelabs.ai}

\author[label17]{Wathsala Herath}
\ead{wathsala.herath@agentsway.ai}

\author[label7]{Chalani Rajapakse}
\ead{chalani.rajapakse@iciclelabs.ai}

\author[label1]{Sachin Shetty}
\ead{sshetty@odu.edu}

\author[label11]{Anita H.\ Clayton}
\ead{AHC8V@uvahealth.org}

\author[label1]{Christopher K.\ Rhea}
\ead{crhea@odu.edu}

\author[label5]{Ng Wee Keong}
\ead{awkng@ntu.edu.sg}

\author[label6]{Kasun De Zoysa}
\ead{kasun@ucsc.cmb.ac.lk}

\author[label8]{Amin Hass}
\ead{amin.hassanzadeh@accenture.com}

\author[label14]{Shaifali Kaushik}
\ead{Skaus2000@gmail.com}

\author[label13]{Preston Samuel}
\ead{preston.l.samuel.mil@health.mil}

\author[label14]{Atmaram Yarlagadda}
\ead{atmaram.yarlagadda.civ@health.mil}


\address[label1]{Old Dominion University, Norfolk, VA, USA}
\address[label10]{AI Motion Labs, Melbourne, Australia}
\address[label5]{Nanyang Technological University, Singapore}
\address[label6]{University of Colombo, Sri Lanka}
\address[label7]{IcicleLabs.AI}
\address[label8]{Accenture Technology Labs, Arlington, VA, USA}
\address[label17]{Agentsway.AI}
\address[label16]{GSI Scandinavia AB}
\address[label11]{Department of Psychiatry and Neurobehavioral Sciences, \\ University of Virginia School of Medicine, Charlottesville, VA, USA}
\address[label13]{Blanchfield Army Community Hospital, Fort Campbell, KY, USA}
\address[label14]{McDonald Army Health Center, Newport News, VA, USA}

\begin{abstract}

Post-Traumatic Stress Disorder (PTSD) is fundamentally a neuroplastic problem — traumatic contact events encode over-reactive neural pathways through Hebbian long-term potentiation, producing hair-triggered amygdala-HPA stress cascades that fire before conscious awareness can intercept them. Existing therapeutic approaches — prolonged exposure, EMDR, cognitive behavioural therapy — operate predominantly downstream of the reactive cascade, teaching patients to tolerate or reframe distress after it has arisen. While clinically valuable, these suppression-based approaches do not produce the upstream pathway dissolution that constitutes lasting structural neural reorganisation. This paper proposes MindGap, a privacy-preserving on-device conversational AI framework that delivers structured neuroplastic rehabilitation for PTSD through the practice of dependent origination — a Buddhist psychological framework that identifies the precise moment between the pre-cognitive affective signal and the reactive elaboration that follows as the site of therapeutic intervention. MindGap guides patients through three progressive layers of observation at this feeling tone gap: noticing the bare affective signal before reactive elaboration, recognising it as self-arising rather than caused by the stimulus, and recognising the conditioned implicit belief beneath the feeling. Each layer corresponds to progressively deeper prefrontal regulatory engagement and progressively deeper long-term depression-mediated weakening of the reactive pathway — producing genuine upstream dissolution rather than downstream suppression. Running entirely on-device with no data egress, MindGap delivers daily calibrated exposure sessions through a fine-tuned lightweight large language model, making it deployable in sensitive clinical and military contexts where cloud-based solutions are not permitted. We present the theoretical framework grounding MindGap in the neuroscience of Hebbian plasticity, prefrontal-amygdala regulation, and experience-dependent neural reorganisation, describe the system architecture and therapeutic protocol, and outline a proposed randomised controlled trial design for empirical validation.

\end{abstract}

\begin{keyword}
Agentic AI \sep Neuroplasticity \sep Neuroscience \sep PTSD \sep Theory of Dependent Origination
\end{keyword}

\end{frontmatter}

\section{Introduction}
\label{sec:introduction}

Post-Traumatic Stress Disorder (PTSD) affects an estimated 20\% of individuals exposed to traumatic events, with particularly high prevalence among military personnel, first responders, and survivors of assault and accident \cite{kessler2005lifetime, bisson2015post}. Its hallmark features — intrusive re-experiencing, hyperarousal, 
avoidance, and negative alterations in cognition and mood — represent not merely psychological distress but a fundamental reorganisation of neural architecture \cite{pitman2012biological, liberzon2015circuits}. The traumatic event encodes an over-reactive threat-response circuit through Hebbian long-term potentiation: the amygdala's rapid pre-cognitive affective evaluation fires at maximum intensity during the traumatic contact event, and the synaptic connections constituting that reactive cascade are co-activated and thereby strengthened \cite{hebb1949organization, mahan2012synaptic}. The result is a brain wired to fire the full threat response — cortisol release, attentional narrowing, behavioral mobilisation — at stimuli that are merely associated with 
the original trauma, operating faster than conscious awareness can intercept \cite{pitman2012biological, arnsten2009stress}. PTSD is, in this precise sense, a neuroplastic encoding problem.

The neuroscience of neuroplasticity establishes that the same Hebbian mechanism that encodes reactive pathways can dissolve them — through long-term synaptic depression of 
pathways that repeatedly fail to complete their activation, and long-term potentiation of the prefrontal regulatory circuits that intercept them \cite{malenka2004ltp, 
citri2008synaptic}. This bidirectionality is the theoretical foundation of existing evidence-based PTSD treatments. Prolonged exposure therapy reduces the emotional response to trauma-related stimuli through repeated activation in the absence of actual danger, gradually weakening the conditioned threat association \cite{foa2007prolonged}. Eye Movement Desensitisation and Reprocessing reduces the intensity of traumatic memory 
through bilateral stimulation during controlled recall \cite{shapiro2001emdr}. Cognitive Behavioural Therapy restructures the negative appraisals and avoidance behaviors that maintain PTSD symptoms \cite{ehlers2000cognitive}. These approaches are clinically effective and represent the current standard of care \cite{bisson2015post, va2017ptsd}.

Yet all three share a fundamental limitation that their clinical efficacy has obscured: they operate primarily \textit{downstream} of the reactive cascade. Prolonged 
exposure activates the full stress response and allows it to habituate — the pathway fires to completion but the feared outcome does not materialise, gradually reducing the conditioned association through extinction learning \cite{foa2007prolonged}. CBT restructures cognitive appraisals after the reactive cascade has already activated — managing the narrative content of the distress rather than intercepting the cascade that 
generated it \cite{ehlers2000cognitive}. In neuroplastic terms, these approaches are suppression-adjacent: the reactive pathway still fires, and while extinction and 
reappraisal produce genuine neural changes, they do not produce the upstream pathway dissolution — the long-term depression-mediated weakening of the reactive pathway before full activation — that constitutes the most direct form of structural neural reorganisation \cite{craske2014maximizing, hebb1949organization}.

The critical intervention point — identified with remarkable precision by the Buddhist psychological theory of dependent origination and confirmed by affective neuroscience — is the moment between the pre-cognitive affective signal and the reactive elaboration that follows it \cite{analayo2003satipatthana, ledoux1996emotional}. This moment — which we term the \textit{feeling tone gap} — corresponds neurobiologically to the prefrontal-amygdala regulatory window: the brief period after amygdala valence tagging in which top-down prefrontal modulation can intercept the downstream propagation of the stress cascade before full HPA activation \cite{ochsner2005cognitive, arnsten2009stress}. When a patient meets the pre-cognitive affective signal at this moment — noticing the bare unpleasantness before it has elaborated into panic, avoidance, or narrative — the reactive pathway does not fire to completion. Repeated across many sessions, this upstream interception weakens the reactive pathway through long-term depression and strengthens the prefrontal observation pathway through long-term potentiation — producing dissolution rather than suppression \cite{hölzel2011mindfulness, 
davidson2012emotional}. This is the neuroplastic mechanism that existing treatments approach indirectly but do not target with precision.

Delivering this intervention in practice requires a therapeutic tool capable of presenting calibrated contact events — stimuli that activate the feeling tone without 
triggering the full reactive cascade — and guiding the patient to observe at the feeling tone gap at the moment of activation. Artificial intelligence agents are uniquely suited to this role~\cite{agentic-workflow-practicle-guide, agentic-ai-transition-organization}. An AI agent is egoless and non-reactive: it cannot inadvertently contaminate the patient's reactive experience through emotional tone, body language, or counter-transference \cite{lucas2014s, woebot2017, astride}. It can present stimuli at precisely 
calibrated intensity levels, adjusted in real time based on patient response. It can deliver high-frequency practice sessions — the neuroplastic frequency advantage that sparse weekly therapy sessions cannot match \cite{deep-psychiatric}. And running entirely on-device with no data egress, it can be deployed in military and clinical contexts where patient data privacy is non-negotiable \cite{va2017ptsd}.

This paper presents \textbf{MindGap} — a privacy-preserving on-device conversational AI framework that delivers structured neuroplastic PTSD rehabilitation through the dependent origination practice framework. MindGap guides patients through three progressive layers of observation at the feeling tone gap: noticing the bare 
affective signal before reactive elaboration (emotional granularity \cite{barrett2017emotions}), recognising the feeling as self-arising rather than caused by the stimulus (decentering \cite{fresco2007initial}), and recognising the conditioned implicit belief beneath the feeling (metacognitive awareness of implicit belief 
\cite{teasdale2002metacognitive}). Each layer corresponds to progressively deeper prefrontal regulatory engagement and progressively deeper long-term depression-mediated 
weakening of the reactive pathway. The system runs as a fine-tuned lightweight large language model on the patient's mobile device — no cloud inference, no data 
transmission, no privacy risk.

This paper makes four primary contributions. First, we establish the neuroplastic theoretical framework grounding MindGap — mapping the dependent origination chain onto 
the neuroscience of PTSD reactive pathway formation and dissolution, and distinguishing the upstream dissolution mechanism from the downstream suppression approaches that 
characterise existing treatments. Second, we present the MindGap system architecture — the fine-tuned on-device LLM, the therapeutic protocol, the stimulus calibration ladder, and the three-layer observation practice framework. Third, we describe the mobile application design and end-to-end patient journey — from clinical referral and intake through daily practice sessions, weekly deep sessions, real-world support, and progress 
tracking. Fourth, we propose a randomised controlled trial design for empirical validation of MindGap's neuroplastic claims — measuring amygdala reactivity, prefrontal-amygdala connectivity, PTSD symptom severity, and feeling tone recognition latency as outcome measures.

The remainder of this paper is structured as follows. Section~\ref{sec:relatedwork} reviews relevant literature across four streams — the neuroscience and treatment of 
PTSD, the theory of dependent origination as a therapeutic framework, AI-assisted mental health interventions, and privacy-preserving on-device AI. Section~\ref{sec:framework} develops the neuroplastic theoretical framework. Section~\ref{sec:system} presents the MindGap system architecture and therapeutic protocol. Section~\ref{sec:app} describes the mobile application and patient journey. Section~\ref{sec:evaluation} proposes the empirical validation design. Section~\ref{sec:conclusion} concludes.

\section{Related Work}
\label{sec:relatedwork}

MindGap sits at the intersection of four bodies of scholarship: the neuroscience and treatment of PTSD, the application of dependent origination as a therapeutic framework, AI-assisted mental health interventions, and privacy-preserving on-device AI. This section reviews each stream, establishing the conceptual foundations of the framework and identifying the gaps that motivate the present work.

\subsection{PTSD: Neuroscience and Existing Treatments}
\label{sec:rw:ptsd}

PTSD is characterised by persistent re-experiencing, hyperarousal, avoidance, and negative alterations in cognition and mood following traumatic exposure \cite{bisson2015post, kessler2005lifetime}. Its neurobiological substrate involves amygdala hyperreactivity, prefrontal-amygdala regulatory disruption, and hypothalamic-pituitary-adrenal axis dysregulation — structural consequences of Hebbian long-term potentiation encoding an over-reactive threat-response circuit during the traumatic event \cite{pitman2012biological, liberzon2015circuits, mahan2012synaptic}. Recovery requires not just symptom management but structural neural reorganisation — the weakening of over-encoded reactive pathways and the strengthening of prefrontal regulatory capacity \cite{hölzel2011mindfulness, davidson2012emotional}.

Current first-line treatments include prolonged exposure therapy \cite{foa2007prolonged}, Eye Movement Desensitisation and Reprocessing \cite{shapiro2001emdr}, cognitive behavioural therapy \cite{ehlers2000cognitive}, and pharmacotherapy \cite{va2017ptsd}. These approaches are clinically effective and well-validated \cite{bisson2015post}. However, they share a structural limitation: all operate primarily downstream of the reactive cascade. Prolonged exposure activates the full stress response and relies on extinction learning — the pathway fires to completion but the feared outcome does not materialise \cite{craske2014maximizing}. CBT restructures cognitive appraisals after the cascade has activated. In neuroplastic terms, these are suppression-adjacent approaches — they produce genuine improvements through extinction and reappraisal, but do not target the upstream feeling tone gap where the reactive cascade can be intercepted before completion, producing long-term depression-mediated pathway dissolution rather than habituated expression \cite{hebb1949organization, malenka2004ltp}.

\subsection{Dependent Origination as a Therapeutic Framework}
\label{sec:rw:dependentorigination}

The Buddhist psychological theory of dependent origination describes a conditioned chain through which sensory contact gives rise to immediate feeling tone — pleasant, unpleasant, or neutral — which conditions craving, aversion, and delusion, producing habitual reactive behavior \cite{analayo2003satipatthana, bodhi2000connected}. Its therapeutic logic lies in the gap between feeling tone and reactive elaboration — the moment the affective signal is present but has not yet precipitated into full reactive cascade — identified as the precise point of possible intervention \cite{analayo2003satipatthana}. This moment corresponds neurobiologically to the prefrontal-amygdala regulatory window, within which top-down modulation can intercept bottom-up amygdala propagation before full HPA activation \cite{ochsner2005cognitive, arnsten2009stress}.

The dialogue between dependent origination and neuroscience has a substantial history. Varela, Thompson, and Rosch argued that Buddhist phenomenology and cognitive science share a commitment to embodied, enacted experience \cite{varela1991embodied}. Brewer and colleagues demonstrated that contemplative practice reduces posterior cingulate cortex activation associated with self-referential craving and reactive elaboration \cite{brewer2011meditation, brewer2013craving}. Teasdale's metacognitive awareness framework — central to mindfulness-based cognitive therapy — operationalises dependent origination's seeing practice in clinical terms: the recognition of thoughts and feelings as mental events rather than facts, associated with durable reduction in depressive relapse \cite{teasdale2002metacognitive}. Fresco and colleagues developed the Experiences Questionnaire to measure decentering — the capacity to observe one's own mental content without identification — as a transdiagnostic therapeutic mechanism \cite{fresco2007initial}. Barrett's research on emotional granularity demonstrates that precise affective labelling reduces amygdala activation and increases prefrontal regulatory engagement \cite{barrett2017emotions}. No existing work, however, has applied the full dependent origination framework specifically to PTSD rehabilitation or proposed it as the therapeutic mechanism for an AI-assisted intervention.

\subsection{AI-Assisted Mental Health Interventions}
\label{sec:rw:aimentalhealth}

Conversational AI systems for mental health have proliferated over the past decade. Woebot delivers CBT-based psychoeducation and mood tracking through a chatbot interface, demonstrating reductions in anxiety and depression symptoms in randomised trials \cite{woebot2017, fitzpatrick2017delivering}. Wysa provides AI-guided emotional support and CBT techniques with reported improvements in wellbeing \cite{inkster2018empathetic}. Replika offers an AI companion with reported benefits for loneliness and social anxiety \cite{skjuve2021my}. In the PTSD domain, the BRAVEMIND virtual reality system delivers prolonged exposure therapy in immersive environments, demonstrating clinical effectiveness in combat veterans \cite{rizzo2010ptsd}. Kognito and similar platforms deliver scenario-based training for mental health skills.

These systems share three limitations relevant to MindGap. First, all operate therapeutically downstream of the reactive cascade — delivering CBT reappraisal, psychoeducation, or habituation-based exposure after the reactive response has already activated, without targeting the upstream feeling tone gap. Second, none employ a neuroplastic dissolution mechanism — the distinction between LTD-mediated pathway weakening through upstream interception and suppression-based downstream management. Third, the majority rely on cloud-based inference, making them unsuitable for deployment in military and sensitive clinical contexts where patient data cannot leave the device \cite{va2017ptsd}.

\subsection{Privacy-Preserving On-Device AI}
\label{sec:rw:ondevice}

The deployment of large language models on mobile devices has become technically feasible with the emergence of efficient small model architectures. Phi-3 Mini \cite{abdin2024phi}, Gemma 2B \cite{team2024gemma}, and LLaMA 3.2 3B \cite{llama-3, llama-4} demonstrate that models in the 2--7 billion parameter range achieve sufficient language understanding and generation quality for structured conversational applications while running within the memory and compute constraints of modern smartphones. Quantisation techniques — 4-bit and 8-bit integer quantisation — further reduce memory footprint without significant quality degradation \cite{dettmers2023qlora, deep-stride}. Framework support for on-device inference has matured through llama.cpp, MLC-LLM, and Apple's Core ML, enabling sub-second response latencies on current flagship devices.

In clinical and military mental health contexts, on-device inference addresses a critical barrier to AI-assisted therapy adoption~\cite{psychiatric-ai-app}. DoD and VA data governance requirements restrict the transmission of patient mental health data to external servers \cite{va2017ptsd}. Existing cloud-based mental health AI systems cannot be deployed in these contexts without significant compliance overhead. On-device inference eliminates this barrier entirely — the model runs locally, session data is stored locally, and no patient information leaves the device. Fine-tuning of small models for specific therapeutic applications has been demonstrated in adjacent domains \cite{yang2024mentallama, nurolense}, though no existing work has fine-tuned an on-device model for dependent origination-based PTSD rehabilitation.

\subsection{Summary and Gap Analysis}
\label{sec:rw:gap}

Table~\ref{tab:relatedwork} summarises how MindGap relates to representative prior work across the four streams reviewed above.

\begin{table*}[!htb]
\centering
\caption{Comparison of MindGap with representative related work. \cmark~=~supported; \xmark~=~not supported; $\sim$~=~partially supported.}
\begin{adjustbox}{width=1\textwidth}
\label{tab:relatedwork}
\begin{tabular}{lcccccccc}
\toprule
\thead{Work} & \thead{PTSD\\focus} & \thead{Upstream\\intervention} & \thead{Feeling tone\\gap} & \thead{Three-layer\\seeing} & \thead{LTD\\dissolution} & \thead{Dependent\\origination} & \thead{On-device\\privacy} & \thead{Mobile\\delivery} \\
\midrule
Foa et al.~\cite{foa2007prolonged} & \cmark & \xmark & \xmark & \xmark & $\sim$ & \xmark & \xmark & \xmark \\
Shapiro~\cite{shapiro2001emdr} & \cmark & \xmark & \xmark & \xmark & $\sim$ & \xmark & \xmark & \xmark \\
Ehlers \& Clark~\cite{ehlers2000cognitive} & \cmark & \xmark & \xmark & \xmark & \xmark & \xmark & \xmark & \xmark \\
Teasdale et al.~\cite{teasdale2002metacognitive} & \xmark & $\sim$ & $\sim$ & $\sim$ & \xmark & $\sim$ & \xmark & \xmark \\
Brewer et al.~\cite{brewer2011meditation} & \xmark & $\sim$ & $\sim$ & \xmark & $\sim$ & $\sim$ & \xmark & \xmark \\
Varela et al.~\cite{varela1991embodied} & \xmark & $\sim$ & \xmark & \xmark & \xmark & $\sim$ & \xmark & \xmark \\
Woebot~\cite{woebot2017} & \xmark & \xmark & \xmark & \xmark & \xmark & \xmark & \xmark & \cmark \\
BRAVEMIND~\cite{rizzo2010ptsd} & \cmark & \xmark & \xmark & \xmark & $\sim$ & \xmark & \xmark & \xmark \\
Fitzpatrick et al.~\cite{fitzpatrick2017delivering} & \xmark & \xmark & \xmark & \xmark & \xmark & \xmark & \xmark & \cmark \\
Yang et al.~\cite{yang2024mentallama} & $\sim$ & \xmark & \xmark & \xmark & \xmark & \xmark & \xmark & \xmark \\
Phi-3 Mini~\cite{abdin2024phi} & \xmark & \xmark & \xmark & \xmark & \xmark & \xmark & \cmark & \cmark \\
Barrett~\cite{barrett2017emotions} & \xmark & $\sim$ & $\sim$ & $\sim$ & $\sim$ & \xmark & \xmark & \xmark \\
\midrule
\textbf{MindGap (proposed)} & \cmark & \cmark & \cmark & \cmark & \cmark & \cmark & \cmark & \cmark \\
\bottomrule
\end{tabular}
\end{adjustbox}
\end{table*}

The table reveals a clear gap. Existing PTSD treatments are clinically grounded but operate downstream and lack AI-assisted delivery. Existing mental health AI systems deliver on mobile but do not employ upstream neuroplastic dissolution mechanisms. Existing contemplative neuroscience work establishes the theoretical convergence between dependent origination and neuroscience but does not operationalise it as a PTSD intervention. Existing on-device AI work provides the technical foundation but has not been applied to therapeutic protocol delivery. MindGap is the first framework to address all eight dimensions simultaneously — combining PTSD clinical focus, upstream feeling tone gap intervention, three-layer seeing practice, LTD dissolution mechanism, dependent origination theoretical grounding, on-device privacy preservation, and mobile delivery.

\section{Theoretical Framework}
\label{sec:framework}

MindGap is grounded in the convergence of three bodies of knowledge: the neuroscience of PTSD as a neuroplastic encoding problem, the Buddhist psychological theory of dependent origination as a phenomenological map of the reactive mind, and the affective neuroscience of prefrontal-amygdala regulation as the mechanism through which upstream intervention produces structural neural reorganisation. This section develops each in turn and establishes the theoretical foundation for the MindGap intervention.

\subsection{PTSD as Neuroplastic Encoding}
\label{sec:framework:ptsd}

PTSD is not simply a psychological response to trauma — it is a physical reorganisation of neural architecture through the mechanism of Hebbian synaptic plasticity. During a traumatic event, the amygdala's pre-cognitive affective evaluation system fires at maximum intensity, generating an extreme unpleasant feeling tone that activates the hypothalamic-pituitary-adrenal stress axis at full force \cite{pitman2012biological, mahan2012synaptic}. Because neurons that fire together wire together \cite{hebb1949organization}, the synaptic connections constituting that reactive cascade are co-activated at extreme intensity and thereby encoded with extraordinary strength through long-term potentiation. The result is not a memory in the ordinary sense but a structural feature of the amygdala's threat-detection circuitry: the activation threshold for the full stress cascade drops, and stimuli that are merely associated with the original trauma — a sound, a smell, a visual pattern — now trigger the complete reactive cascade at an intensity disproportionate to any objective threat they represent \cite{liberzon2015circuits, arnsten2009stress}.

\begin{figure}[H]
\centering
\includegraphics[width=\textwidth]{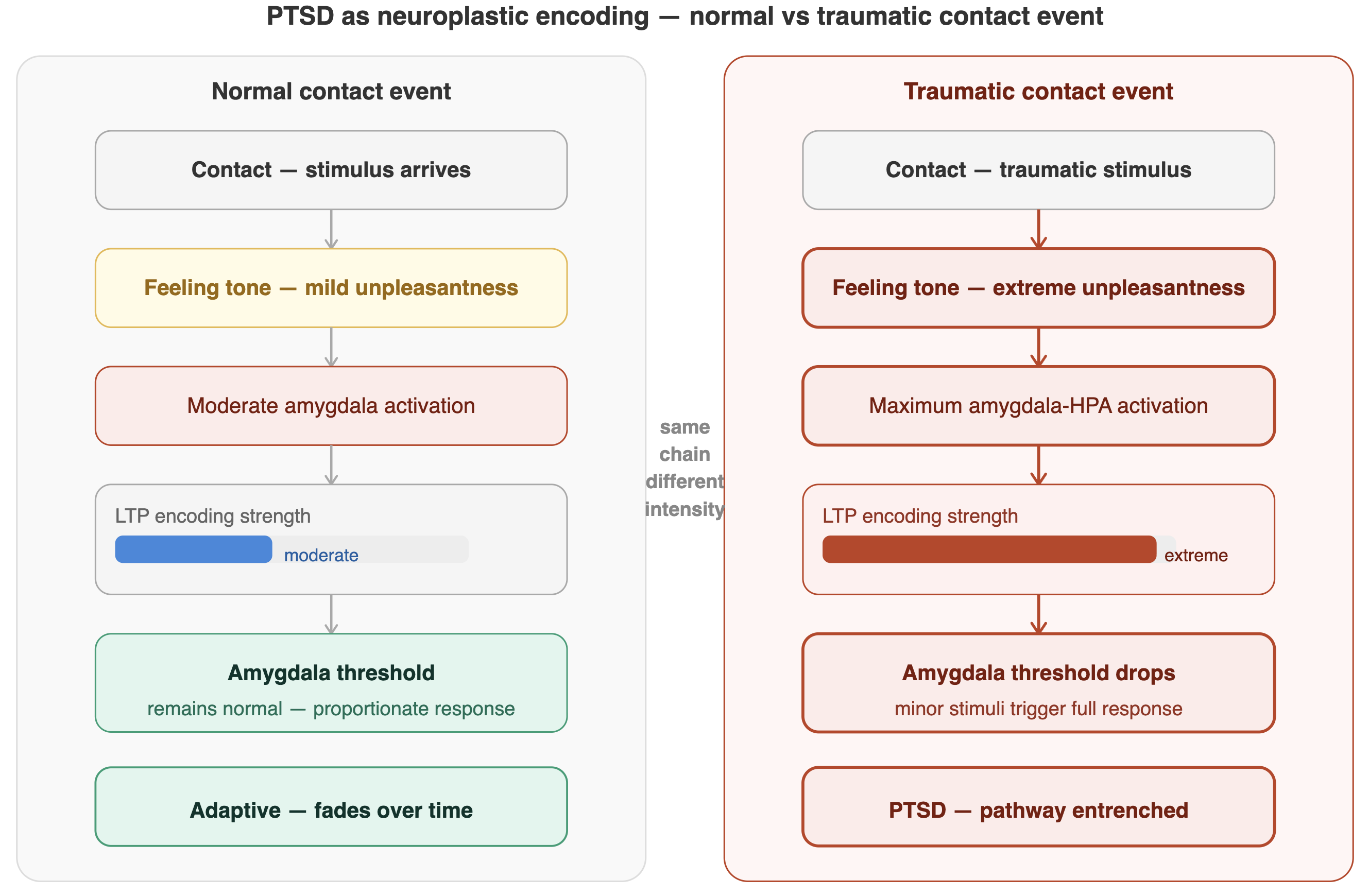}
\caption{PTSD as neuroplastic encoding. Both normal and traumatic contact events follow the same dependent origination chain — contact, feeling tone, amygdala activation, reactive behavior. The difference is quantitative: the traumatic event produces extreme LTP encoding, dropping the amygdala threshold so that minor stimuli subsequently trigger the full reactive cascade. PTSD is the structural consequence of this extreme encoding rather than a distinct psychological process.}
\label{fig:ptsdencoding}
\end{figure}

This threshold drop explains the hallmark features of PTSD. Hyperarousal — the persistent sense of being on edge — reflects a chronically lowered amygdala activation threshold. Re-experiencing — flashbacks and intrusive memories — reflects the reactive pathway firing spontaneously when associated stimuli are encountered. Avoidance reflects the behavioral consequence of a system that has learned to minimise contact with activating stimuli \cite{pitman2012biological}. Each episode of activation in the absence of effective therapeutic intervention reinforces the pathway further through long-term potentiation — the conditioned formation deepens \cite{mahan2012synaptic, citri2008synaptic}. PTSD does not maintain itself through psychological processes alone. It maintains itself through repeated neuroplastic reinforcement of an over-encoded reactive circuit. Figure~\ref{fig:ptsdencoding} illustrates how the same dependent origination chain that operates in ordinary contact events produces extreme encoding in traumatic ones.

\subsection{The Feeling Tone Gap as the Upstream Intervention Point}
\label{sec:framework:feelingtonegap}

The Buddhist psychological theory of dependent origination identifies with remarkable precision the moment in the reactive sequence where intervention is both possible and neuroplastically consequential. The chain proceeds: contact — the stimulus arriving in awareness; feeling tone — the immediate pre-cognitive affective signal of pleasant, unpleasant, or neutral; craving or aversion — the reactive elaboration that follows feeling tone; delusion — the narrative distortion that amplifies and justifies the reaction; conditioned formation — the habitual pattern deepened by each unconscious cycle \cite{analayo2003satipatthana, bodhi2000connected}. The gap between feeling tone and the arising of craving or aversion — the moment the affective signal is present but has not yet precipitated into full reactive elaboration — is identified as the precise site of possible intervention \cite{analayo2003satipatthana}.

Affective neuroscience has independently identified the same moment. The amygdala's rapid valence tagging operates within 100--200 milliseconds of stimulus onset, prior to and independent of conscious cortical appraisal \cite{ledoux1996emotional, zajonc1980feeling}. But the prefrontal cortex, when sufficiently engaged, can modulate the amygdala's downstream propagation — activating top-down regulatory inhibition that reduces the intensity of the HPA cascade before it fully mobilises \cite{ochsner2005cognitive, arnsten2009stress}. This prefrontal-amygdala regulatory window corresponds precisely to the feeling tone gap: the brief period after affective signal generation in which conscious engagement can intercept the downstream cascade. When a patient meets the feeling tone — the bare unpleasantness arising at the moment of stimulus contact — with observing awareness rather than reactive elaboration, the prefrontal cortex activates, the cascade is modulated before full HPA activation, and the reactive pathway fails to complete its activation. A pathway that does not fire to completion begins to weaken through long-term synaptic depression \cite{malenka2004ltp, citri2008synaptic}.

\begin{figure}[H]
\centering
\includegraphics[width=\textwidth]{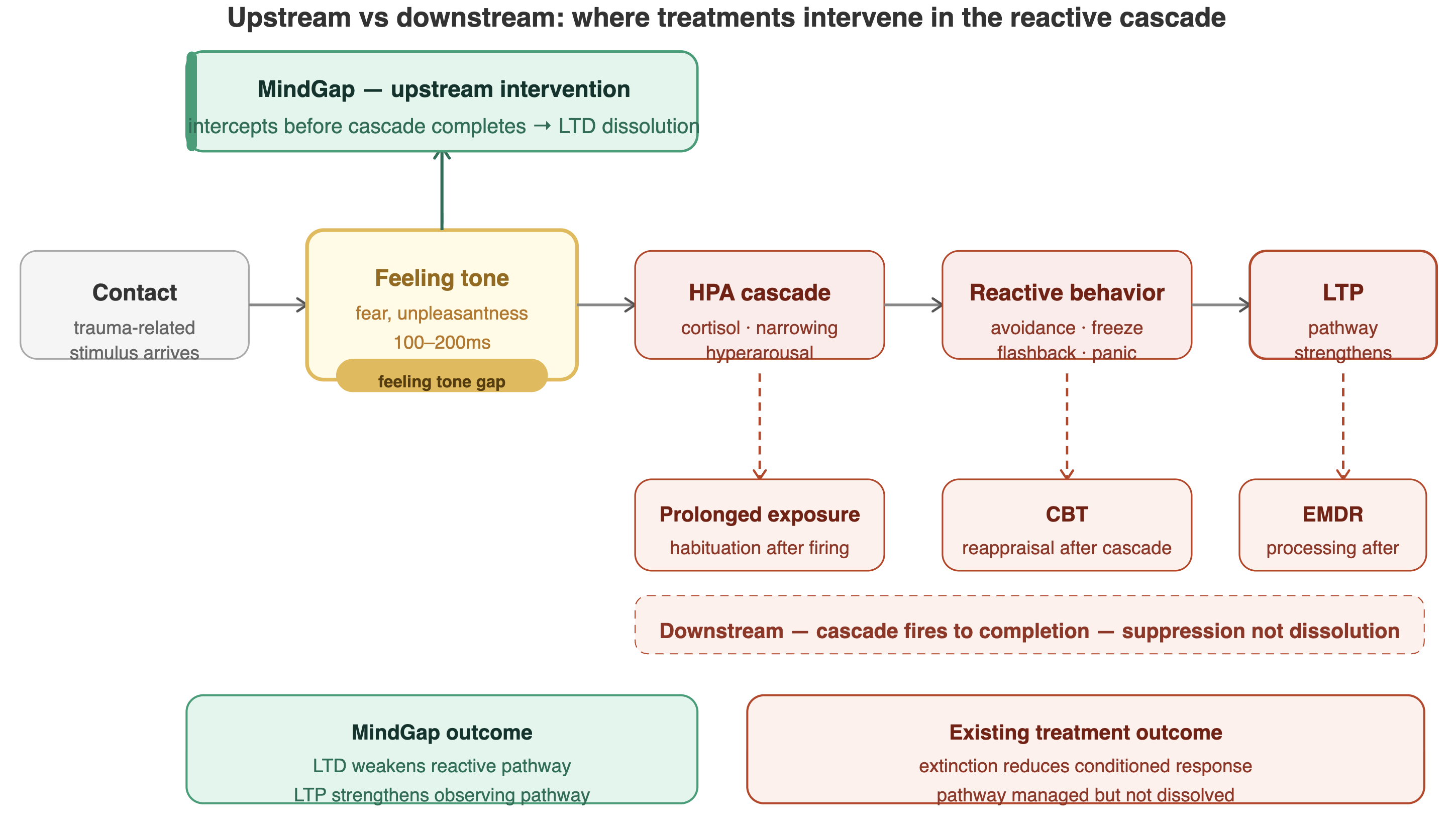}
\caption{Upstream vs downstream intervention in the PTSD reactive cascade. MindGap intervenes at the feeling tone gap — before the HPA cascade fires to completion — producing LTD-mediated weakening of the reactive pathway. Existing treatments — prolonged exposure, CBT, EMDR — intervene downstream, after the cascade has activated, producing clinical benefits through extinction and reappraisal but not through upstream pathway dissolution.}
\label{fig:upstreamdownstream}
\end{figure}

This upstream interception is neuroplastically distinct from the mechanisms of existing PTSD treatments, as illustrated in Figure~\ref{fig:upstreamdownstream}. Prolonged exposure therapy activates the full stress response and relies on extinction learning — the pathway fires to completion but the feared outcome does not materialise \cite{foa2007prolonged, craske2014maximizing}. Cognitive behavioural therapy restructures appraisals after the cascade has activated \cite{ehlers2000cognitive}. EMDR processes traumatic memory after the reactive cascade has fired \cite{shapiro2001emdr}. All three are clinically effective through downstream mechanisms. None produces the upstream LTD-mediated dissolution that constitutes the most direct form of reactive pathway weakening — because none targets the feeling tone gap where the cascade can be intercepted before completion.

\subsection{The Three Layers of Therapeutic Seeing}
\label{sec:framework:threelayers}

The practice of meeting feeling tone with awareness at the feeling tone gap operates at three progressive depths, each corresponding to a deeper level of prefrontal regulatory engagement and a more durable form of LTD-mediated pathway weakening \cite{teasdale2002metacognitive, barrett2017emotions, fresco2007initial}. These three layers constitute the therapeutic protocol through which MindGap guides patients.

\textbf{Layer 1 — Emotional granularity.} The most accessible layer is the precise labelling of the affective signal before reactive elaboration begins. Rather than being swept from feeling tone into panic or avoidance automatically, the patient simply notices and names: \textit{there is fear here} or \textit{there is unpleasantness here}. Barrett's research on emotional granularity demonstrates that this precise affective labelling activates the prefrontal cortex's regulatory engagement with the amygdala, reducing amygdala activation intensity and beginning to modulate the downstream cascade \cite{barrett2017emotions, proof-of-tbi}. In neuroplastic terms, this is the first intervention: the reactive pathway does not fire at full intensity, and the observing pathway — the prefrontal circuit that labelled the feeling — is activated and begins to strengthen through long-term potentiation.

\textbf{Layer 2 — Decentering.} The second layer involves recognising the feeling as a mental event arising in the patient's mind rather than as a property of the external stimulus. The patient recognises: \textit{this fear is arising in me — it is not caused by the sound}. This decentering \cite{fresco2007initial} — the capacity to observe one's own mental content as mental content rather than as objective reality — reduces default mode network narrative capture and is associated with more durable attenuation of reactive patterns than affective labelling alone \cite{teasdale2002metacognitive}. For PTSD patients, decentering is the recognition that the traumatic stimulus is not inherently dangerous — it is the conditioned reactive pathway, not the objective threat level, that is generating the distress.

\textbf{Layer 3 — Metacognitive awareness of implicit belief.} The deepest layer involves recognising the conditioned implicit belief beneath the feeling — the gathi, in dependent origination terms — as a conditioned formation rather than a truth. The patient recognises: \textit{this fear arises because I carry the implicit belief that sudden sounds mean mortal danger — a belief formed during trauma, not an accurate perception of the present situation}. Teasdale's research identifies this metacognitive awareness of implicit belief as the layer of contemplative practice most strongly associated with durable reduction in reactive relapse, because it addresses the conditioned formation itself rather than its immediate expression \cite{teasdale2002metacognitive}. In neuroplastic terms, this is where the gathi begins to dissolve — the implicit belief loses its automatic grip on the amygdala's threat-evaluation system, and the stimulus loses its power to trigger the full reactive cascade.

\subsection{Why AI Agents Enable This Intervention}
\label{sec:framework:whyai}

Delivering the feeling tone gap intervention clinically requires a tool that can present calibrated contact events at controlled intensity, prompt the patient to observe at the feeling tone gap at the moment of activation, and sustain this practice at the high frequency that neuroplastic dissolution requires. Human therapists are valuable for this role but structurally constrained: session frequency is limited, the therapist's own emotional presence can inadvertently contaminate the patient's reactive experience, and the therapeutic relationship introduces social dynamics that complicate the pure observation of self-arising reactivity \cite{lucas2014s}.

AI agents address all three constraints. An AI agent is egoless — it has no emotional investment in the interaction, no counter-transference, no body language or vocal tone that could contaminate the patient's reactive experience~\cite{towards-rai-xai}. Every feeling arising in the patient during an AI agent session is demonstrably self-arising, providing the clean practice conditions that the feeling tone gap intervention requires. An AI agent can deliver sessions daily — at the frequency the neuroplastic work demands — without fatigue, frustration, or scheduling constraints. And running on-device with no data egress, it can provide this daily high-frequency practice in the patient's own environment — at the moment of real-world activation, not only in the clinical session — extending the neuroplastic training to the contexts where it is most needed.

\subsection{Dissolution Versus Suppression: The Neuroplastic Distinction}
\label{sec:framework:dissolution}

The neuroplastic mechanism underlying MindGap's therapeutic protocol is dissolution — upstream interception of the reactive cascade at the feeling tone gap, producing long-term depression-mediated weakening of the reactive pathway — which is structurally distinct from the suppression that characterises most existing PTSD interventions. Suppression is the management of reactive output after the cascade has fired: the patient experiences the full stress response but learns to tolerate it, to reframe it, or to prevent its behavioral expression. In neuroplastic terms, suppression maintains the reactive pathway — it fires to completion and is strengthened through long-term potentiation each time, regardless of whether behavioral expression is managed \cite{hebb1949organization, citri2008synaptic}.

Dissolution occurs when the feeling tone gap intervention successfully intercepts the cascade before completion. The amygdala has signalled, but the prefrontal observing engagement has modulated the downstream propagation before full HPA activation. The reactive pathway has not fired to completion. Repeated across many sessions, this non-completion weakens the pathway through long-term depression — the synaptic equivalent of a conditioned circuit that is systematically deprived of the co-activation it needs to be maintained \cite{malenka2004ltp}. Simultaneously, the prefrontal observing pathway that activated to intercept the cascade strengthens through long-term potentiation. The balance of the prefrontal-amygdala regulatory relationship shifts — the threshold rises, the cascade fires less forcefully, and the patient's capacity for upstream interception grows more automatic and accessible over time \cite{hölzel2011mindfulness, davidson2012emotional}. This is the neuroplastic mechanism that MindGap is designed to operationalise — and that distinguishes it from the clinically valuable but suppression-based approaches that constitute the current standard of PTSD care.

\subsection{The Dependent Origination Chain in PTSD}
\label{sec:framework:chain}

\begin{figure}[H]
\centering
\includegraphics[width=\textwidth]{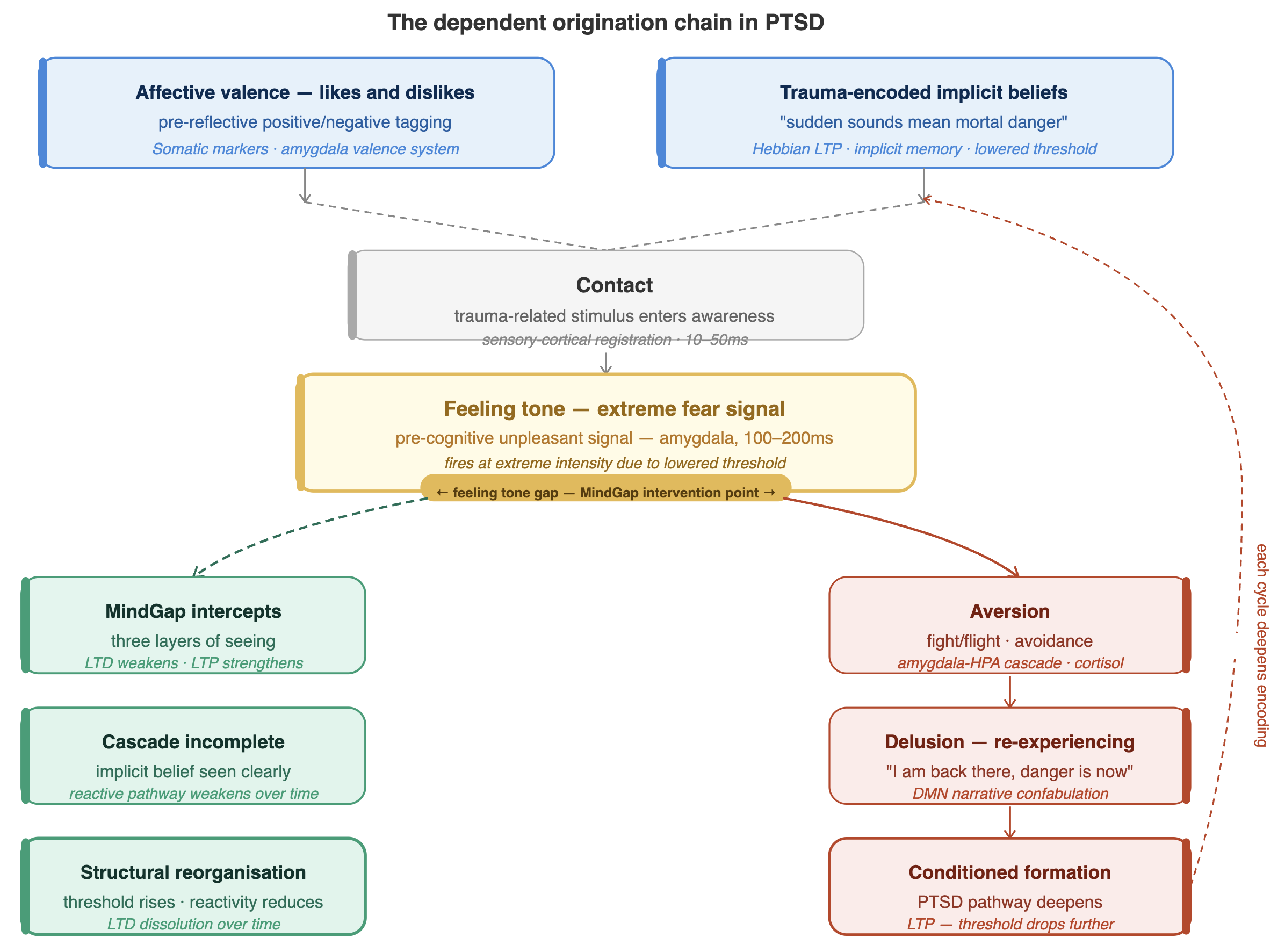}
\caption{The dependent origination chain in PTSD. Trauma-encoded implicit beliefs and affective valence form the conditioned substrate. At the feeling tone gap the chain diverges: MindGap's three-layer seeing practice intercepts the cascade upstream (green, left), weakening the reactive pathway through long-term depression and strengthening the observing pathway through long-term potentiation. The unchecked reactive path (red, right) proceeds through aversion, re-experiencing delusion, and conditioned formation deepening — each cycle lowering the amygdala threshold further through Hebbian long-term potentiation.}
\label{fig:ptsdchain}
\end{figure}

The theory of dependent origination provides a phenomenological map of the reactive mind precise enough to identify not just that PTSD reactive responses arise but exactly how they arise, link by link, and at precisely which point in the sequence conscious awareness can intervene \cite{analayo2003satipatthana, bodhi2000connected}. Figure~\ref{fig:ptsdchain} illustrates the full chain as it operates in the PTSD context.

Two layers of conditioned substrate precede every contact event. The first is \textit{affective valence} — the brain's automatic positive/negative tagging system, operating continuously and pre-reflectively below conscious awareness through somatic markers and the amygdala's valence evaluation circuitry \cite{damasio1994descartes, ledoux1996emotional}. The second — and more clinically significant in PTSD — is \textit{trauma-encoded implicit belief}: the deep conditioned threat-response expectations formed during the traumatic event and encoded below conscious awareness through extreme Hebbian long-term potentiation \cite{mahan2012synaptic, pitman2012biological}. These implicit beliefs — \textit{sudden loud sounds mean mortal danger; this environment means I am under threat; vulnerability means annihilation} — function not as conscious cognitions but as structural features of the amygdala's threat-evaluation system, lowering its activation threshold for entire classes of stimuli associated with the original trauma. They are the neurobiological equivalent of what dependent origination calls deeply conditioned formation — the implicit substrate from which the reactive cascade arises \cite{analayo2003satipatthana}.

\textbf{Contact} occurs when a trauma-related stimulus enters awareness. At the sensory-cortical level this is a neutral event — the stimulus is registered before any affective quality has been assigned \cite{kandel2013principles}. But contact with a stimulus that the amygdala's implicit threat-evaluation system has been conditioned to recognise as danger immediately triggers the next link.

\textbf{Feeling tone} arises pre-cognitively — the amygdala's rapid valence evaluation tags the stimulus as extremely unpleasant within 100--200 milliseconds, prior to and independent of conscious cortical appraisal \cite{ledoux1996emotional, zajonc1980feeling}. In PTSD this signal fires at extreme intensity because the amygdala threshold has been structurally lowered by the original traumatic encoding. This is not yet fear in the full emotional sense — it is the pre-reflective affective seed from which the full reactive cascade will unfold. It is also the feeling tone gap — the brief window before craving or aversion has fully mobilised — and the precise point at which MindGap intervenes.

\textbf{Aversion} arises automatically from the extreme unpleasant feeling tone — the mind's conditioned movement away from the aversive signal, generating the impulse to escape, avoid, or eliminate the stimulus \cite{pitman2012biological, arnsten2009stress}. In PTSD this manifests as hyperarousal, fight-or-flight activation, and the avoidance behaviors that maintain the disorder. The aversion is not a deliberate choice — it is the automatic conditioned response of a threat-evaluation system that has learned through traumatic encoding that this class of stimulus means mortal danger. The amygdala-HPA stress axis activates, cortisol is released, attentional scope narrows, and the organism mobilises for survival.

\textbf{Delusion} — the narrative distortion arising in the wake of aversion — manifests in PTSD as the compelling re-experiencing narrative: \textit{I am back there, this is happening again, I am not safe right now} \cite{raichle2001default, brewer2013craving}. The default mode network constructs a present-tense account around the feeling tone that mistakes a conditioned neural response for an accurate perception of current reality. This is the mechanism of flashback and intrusive re-experiencing — not a memory playback but a present-tense narrative generated by the reactive cascade meeting the default mode network's narrative-construction processes.

\textbf{Conditioned formation} deepens with each cycle that runs without conscious upstream interception \cite{hebb1949organization, citri2008synaptic}. Every PTSD episode that completes the full reactive cascade — contact, extreme feeling tone, aversion, hyperarousal, re-experiencing narrative — co-activates and thereby strengthens through Hebbian long-term potentiation the synaptic pathway of that cascade. The amygdala threshold drops further. The next episode fires faster and at lower stimulus intensity. The trauma-encoded implicit beliefs deepen — the conditioned formation becomes more entrenched, not less, with each unchecked reactive cycle.

This chain makes precise what the neurobiological account of PTSD alone cannot: it identifies the feeling tone gap — between feeling tone and the arising of aversion — as the moment where the cascade can be intercepted before completion, and where MindGap's three-layer therapeutic protocol operates. It also identifies the deepest therapeutic target: not the reactive behavior, not the emotional response, but the trauma-encoded implicit beliefs that generate the extreme feeling tone in the first place. Layer 3 of the MindGap practice — metacognitive awareness of implicit belief — is specifically designed to make these beliefs visible as conditioned formations rather than accurate threat perceptions, at the moment of feeling tone, before the cascade has run to completion.

\section{System Architecture}
\label{sec:system}

\begin{figure}[H]
\centering
\includegraphics[width=\textwidth]{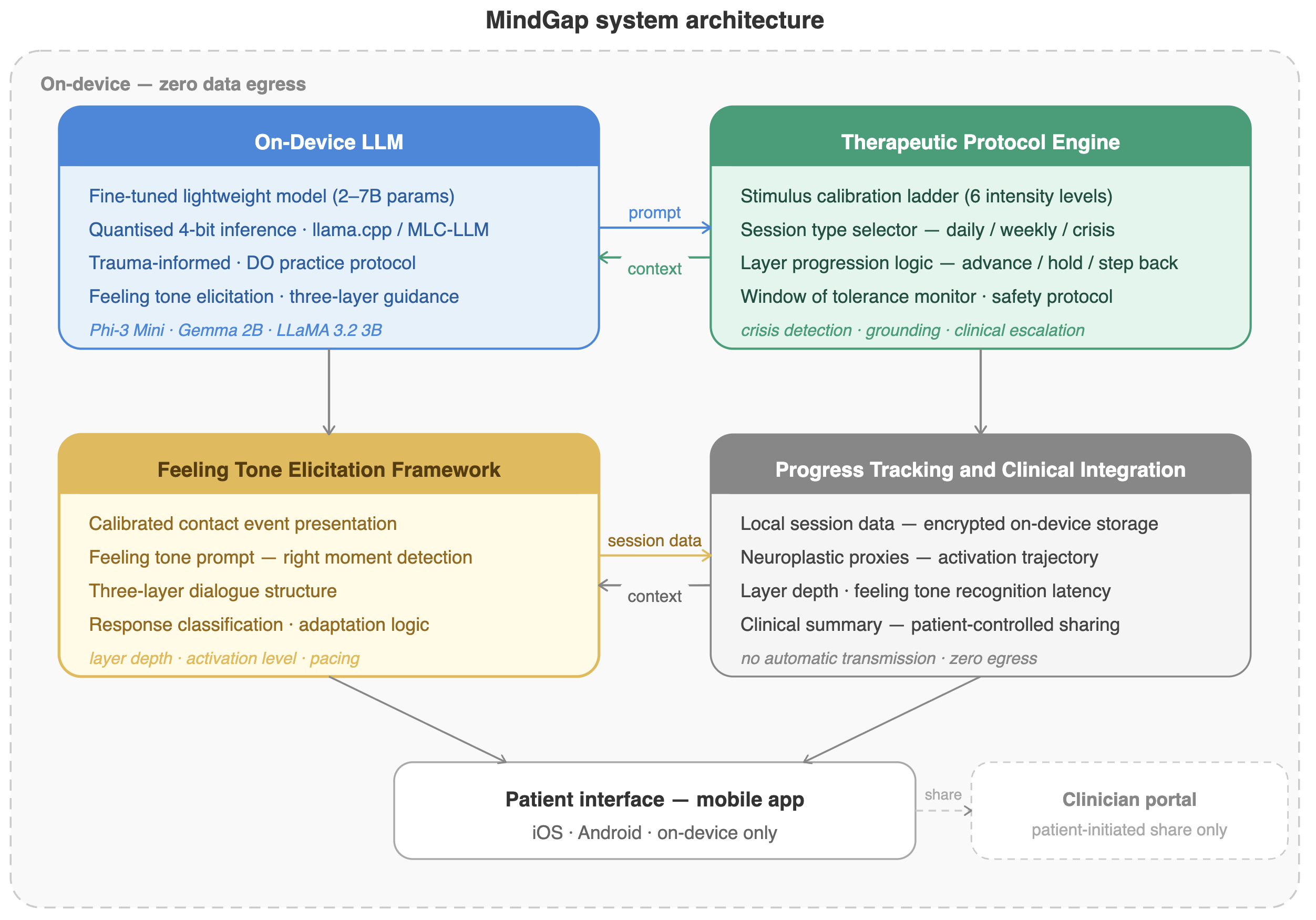}
\caption{MindGap system architecture. All four components run on-device with zero data egress. The on-device LLM and therapeutic protocol engine exchange prompts and session context bidirectionally. Session data flows to the progress tracking module. The feeling tone elicitation framework mediates between the LLM and the protocol engine, translating protocol decisions into therapeutic dialogue. Clinical data sharing is patient-initiated only — no automatic transmission occurs.}
\label{fig:architecture}
\end{figure}

MindGap comprises four tightly integrated components, all running on the patient's mobile device with no data egress: an on-device fine-tuned large language model, a therapeutic protocol engine, a feeling tone elicitation framework, and a progress tracking and clinical integration module. Figure~\ref{fig:architecture} presents the complete system architecture and the data flows between components. This section describes each component in turn.

\subsection{On-Device LLM Architecture}
\label{sec:system:llm}

The foundation of MindGap is a fine-tuned lightweight large language model running entirely on the patient's mobile device. The on-device constraint is not a technical compromise but a clinical and ethical requirement: military personnel and sensitive clinical populations cannot transmit mental health session data to external servers, and existing cloud-based mental health AI systems are therefore unsuitable for these deployment contexts \cite{va2017ptsd}. On-device inference eliminates this barrier entirely — the model runs locally, session data is stored locally, and no patient information leaves the device at any point.

Lightweight models in the 2--7 billion parameter range are technically sufficient for the MindGap therapeutic task. Unlike general-purpose conversational AI applications that require broad world knowledge and creative generation, MindGap requires a bounded set of capabilities: presenting calibrated contact event stimuli in natural language, prompting feeling tone observation at the correct moment, guiding the patient through three structured layers of seeing, detecting activation levels from patient responses, and adapting session pacing accordingly. Candidate architectures include Phi-3 Mini \cite{abdin2024phi}, Gemma 2B \cite{team2024gemma}, and LLaMA 3.2 3B \cite{llama-3, llama-4}, all of which demonstrate sufficient language understanding and generation quality for structured therapeutic dialogue while operating within the memory and compute constraints of current flagship smartphones. Four-bit integer quantisation through QLoRA-compatible frameworks \cite{dettmers2023qlora} reduces memory footprint without significant quality degradation for the structured dialogue task MindGap requires. Inference is managed through llama.cpp, MLC-LLM, or platform-native frameworks such as Apple Core ML, targeting response latencies below two seconds on current flagship devices.

The model is fine-tuned on three corpora. The first is a structured corpus of dependent origination practice dialogues — therapeutic conversations demonstrating correct feeling tone elicitation, three-layer guidance, and adaptive response to patient activation levels. The second is a trauma-informed therapeutic language corpus, ensuring the model's tone, pacing, and vocabulary are appropriate for PTSD populations and consistent with trauma-informed care principles. The third is a feeling tone elicitation corpus — structured examples of the specific prompting pattern that guides patients from bare affective labelling through decentering to metacognitive recognition of implicit belief. Together these three fine-tuning corpora produce a model that is not broadly therapeutically capable but is precisely calibrated for the MindGap intervention — a narrow, deep capability rather than broad general therapeutic intelligence.

\subsection{Therapeutic Protocol Engine}
\label{sec:system:protocol}

The therapeutic protocol engine is the clinical logic layer of MindGap — the component that determines what happens in each session, at what intensity, and with what therapeutic objective. It operates as a state machine that maintains the patient's current position in the therapeutic trajectory and makes real-time decisions about stimulus intensity, layer depth, pacing, and safety response.

\begin{figure}[H]
\centering
\includegraphics[width=\textwidth]{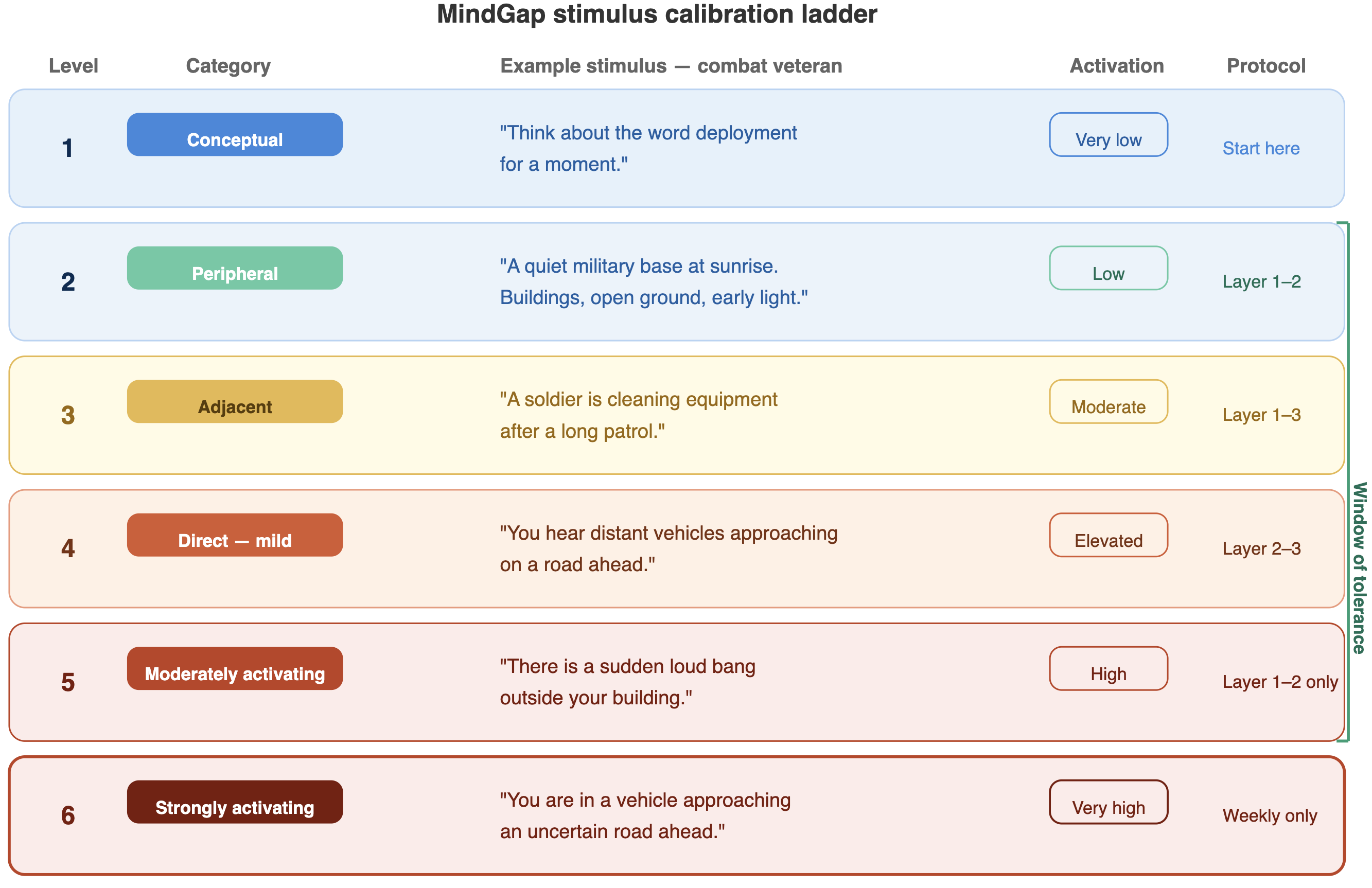}
\caption{MindGap stimulus calibration ladder. Six intensity levels progress from conceptual and abstract (Level 1) through peripheral, adjacent, and direct stimuli to strongly activating (Level 6). Example stimuli are shown for a combat veteran with primary triggers around vehicles and sudden loud sounds. Levels 2--5 define the therapeutic window of tolerance — the zone within which the patient can observe without being overwhelmed and within which neuroplastic dissolution occurs. The agent advances the ladder only when stable Layer 1 observation is demonstrated at the current level across three consecutive sessions.}
\label{fig:ladder}
\end{figure}

The core of the protocol engine is the stimulus calibration ladder — a patient-specific six-level intensity hierarchy constructed during intake from the patient's trauma profile, primary triggers, current symptom severity, and avoidance patterns. Figure~\ref{fig:ladder} illustrates the ladder structure with example stimuli for a combat veteran. Level 1 stimuli are conceptual and abstract — a single word or general concept associated with the trauma domain — producing minimal activation in most patients. Levels 2 and 3 are peripheral and adjacent — scenarios that relate to the trauma context without directly representing it. Levels 4 and 5 are direct and moderately activating — scenarios that more closely approximate the trigger context. Level 6 is strongly activating — the most intense stimuli reserved for weekly deep sessions after the patient has demonstrated stable three-layer observation at lower levels. Levels 2 through 5 define the therapeutic window of tolerance — the activation zone within which the patient can observe without being overwhelmed and within which genuine neuroplastic work occurs \cite{ogden2006trauma}.

The engine selects stimulus level and session type — daily practice (5--10 minutes), weekly deep session (30--45 minutes), or real-world support (unscheduled) — based on the patient's current ladder position, the time since the last session, the activation level reported at check-in, and the layer depth achieved in recent sessions. Ladder advancement follows a conservative criterion: the patient must demonstrate stable Layer 1 observation at the current level across three consecutive sessions before the engine advances to the next level. Regression — stepping back to a lower level — occurs automatically when activation exceeds the window of tolerance threshold or when the patient reports distress above a self-report threshold during the session.

The safety protocol is a critical component of the engine. When patient responses indicate activation approaching or exceeding the upper boundary of the window of tolerance, the engine immediately suspends the contact event sequence, delivers a structured grounding protocol — orienting to immediate sensory environment, slowing breath, naming five present-moment observations — and steps back to a lower stimulus level. When responses indicate acute distress — activation reports at maximum intensity, explicit expressions of crisis, or response patterns consistent with dissociation — the engine transitions to crisis support mode, provides immediate stabilisation support, and presents a direct pathway to clinical contact or crisis resources. The safety protocol is designed to ensure MindGap never functions as the sole support for a patient in acute distress, consistently positioning itself as a complement to rather than a replacement for clinical care \cite{va2017ptsd}.

\subsection{Feeling Tone Elicitation Framework}
\label{sec:system:elicitation}

\begin{figure}[H]
\centering
\includegraphics[width=\textwidth]{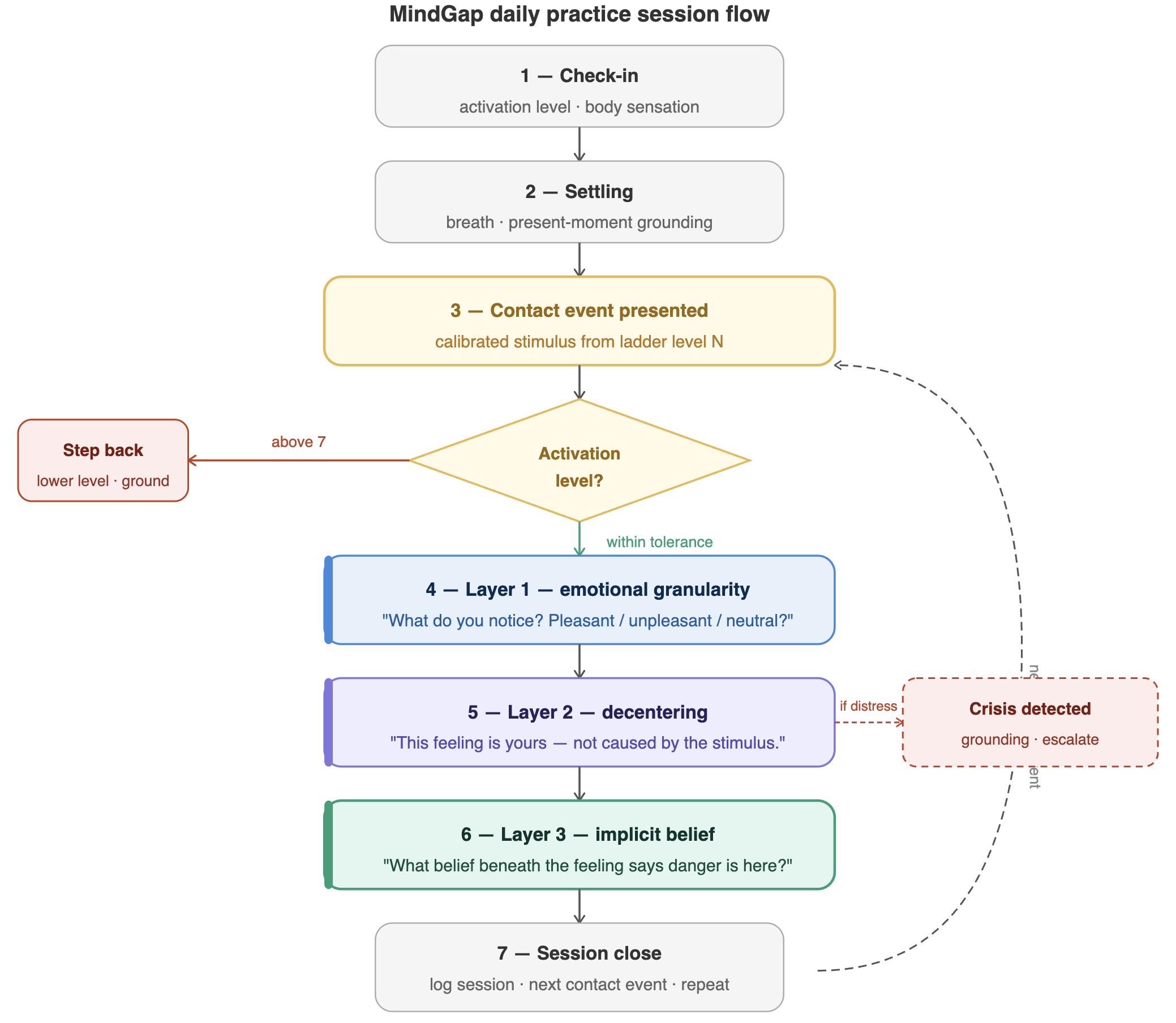}
\caption{MindGap daily practice session flow. Each session proceeds from check-in and settling through contact event presentation and activation assessment. Within the therapeutic window of tolerance, the agent guides the patient through three layers of seeing — emotional granularity (Layer 1), decentering (Layer 2), and metacognitive awareness of implicit belief (Layer 3). Sessions repeat across multiple contact events. Crisis detection triggers immediate grounding and clinical escalation. The step-back path activates automatically when activation exceeds the window of tolerance threshold.}
\label{fig:sessionflow}
\end{figure}

The feeling tone elicitation framework is the therapeutic core of MindGap — the component that mediates between the protocol engine's clinical decisions and the patient's moment-to-moment experience. It translates the abstract structure of the three-layer seeing practice into concrete therapeutic dialogue, managing the timing, phrasing, and sequencing of prompts that guide the patient to observe at the feeling tone gap. Figure~\ref{fig:sessionflow} illustrates the complete session flow including the elicitation framework's decision points.

Contact event presentation follows a structured format designed to maximise the clarity of the feeling tone signal while minimising narrative elaboration that would bypass the feeling tone gap. Stimuli are presented in second-person present tense — \textit{"you are in a vehicle approaching a road ahead"} — rather than third-person narrative, to maximise ecological validity and amygdala engagement. A brief pause is built into the prompt structure after stimulus presentation, allowing the affective signal to arise before any evaluative prompt is introduced.

The feeling tone prompt is delivered at a specific moment — after the stimulus has been presented and before the patient has had time to form a full cognitive appraisal. The prompt is deliberately minimal: \textit{"What do you notice right now? Is there a feeling of pleasant, unpleasant, or neutral?"} This minimalism is therapeutically intentional — elaborated prompts risk directing the patient's attention to cognitive content rather than the bare affective signal, bypassing Layer 1 and reducing the neuroplastic precision of the intervention.

Three-layer guidance proceeds sequentially from the patient's Layer 1 response. The transition from Layer 1 to Layer 2 is made when the patient has successfully labelled the bare affective signal — the agent confirms the label and introduces the decentering reframe: \textit{"That feeling is arising in you right now. The scenario is just words — the feeling belongs to you, not to it. Can you notice that?"}. The transition to Layer 3 is made more carefully — only when the patient has demonstrated stable Layer 2 observation, and only with a gentle, open prompt that does not suggest the content of the belief: \textit{"What does this feeling seem to believe about this situation? If the feeling could speak, what would it say?"}. This open formulation is critical — directing the patient toward a specific belief content risks cognitive override of the genuine implicit belief recognition that Layer 3 targets.

Response classification operates on patient outputs in real time, categorising responses along two dimensions: layer depth reached (1, 2, or 3) and activation level (within tolerance, approaching upper boundary, or exceeding upper boundary). These classifications drive the engine's subsequent decisions — whether to proceed to the next layer, hold at the current layer, step back to a lower stimulus level, or activate the safety protocol. Response classification is performed by the fine-tuned LLM using a structured output format, ensuring consistent classification across sessions and patients.

\subsection{Progress Tracking and Clinical Integration}
\label{sec:system:tracking}

All session data is stored locally on the patient's device in an encrypted data store, with no automatic transmission to any external system. The progress tracking module maintains a longitudinal record of session outcomes and computes neuroplastic progress proxies — behavioural biomarkers that serve as on-device approximations of neuroplastic change without requiring neuroimaging.

Three primary neuroplastic proxies are tracked. The first is \textit{activation trajectory} — the trend in self-reported activation levels at the start and end of sessions over time. Reducing activation levels across equivalent stimulus intensities is consistent with the predicted LTD-mediated weakening of the reactive pathway and the rising amygdala threshold that sustained MindGap practice should produce \cite{hölzel2011mindfulness, davidson2012emotional}. The second is \textit{layer depth progression} — the proportion of sessions in which the patient reaches Layer 2 and Layer 3 over time. Increasing layer depth access is consistent with the predicted LTP-mediated strengthening of the prefrontal observing pathway, which should make deeper observation more automatic and accessible as practice continues \cite{teasdale2002metacognitive}. The third is \textit{feeling tone recognition latency} — the time between contact event presentation and the patient's feeling tone label, estimated from session interaction timing. Decreasing latency is consistent with the predicted increase in the speed and automaticity of the observing pathway relative to the reactive cascade.

The progress tracking module generates a structured session summary on a monthly basis — a concise report of session frequency, activation trajectory, layer depth progression, stimulus level reached, and the stimulus categories producing highest activation. This summary is presented to the patient within the app and can be shared with the clinical team through a patient-initiated export — a deliberate design decision that maintains patient agency over clinical data sharing and ensures no information leaves the device without explicit patient action. The clinical team uses the summary to inform ladder progression decisions, identify stimulus categories requiring focused attention in formal therapy sessions, and assess whether adjunctive pharmacotherapy or therapy intensity adjustment may be warranted.

\section{Mobile Application and Patient Journey}
\label{sec:app}

The MindGap mobile application delivers the therapeutic protocol described in Section~\ref{sec:system} through a patient-facing interface designed around three principles: clinical accessibility, therapeutic precision, and unconditional privacy. Every interaction is local — no data leaves the device, no network connection is required for session delivery, and the patient retains complete control over what, if anything, is shared with their clinical team~\cite{aitrust-os}. This section traces the complete patient journey from initial onboarding through daily practice, real-world support, weekly deep sessions, and longitudinal progress review. Figure~\ref{fig:screens:onboarding}, Figure~\ref{fig:screens:practice}, and Figure~\ref{fig:screens:progress} illustrate the application interface at each stage.

\subsection{Onboarding and Intake}
\label{sec:app:onboarding}

The patient enters MindGap through a clinician-issued setup code generated during the clinical referral process. The setup code links the patient's app instance to their clinical profile — establishing the therapeutic context within which the stimulus ladder will be constructed — without transmitting any personal health information. The patient downloads the application, enters the setup code, and is greeted by a brief orientation that establishes two commitments central to the therapeutic relationship: that everything stays on the device, and that the patient is always in control of the pace.

\begin{figure}[H]
\centering
\includegraphics[width=0.85\textwidth]{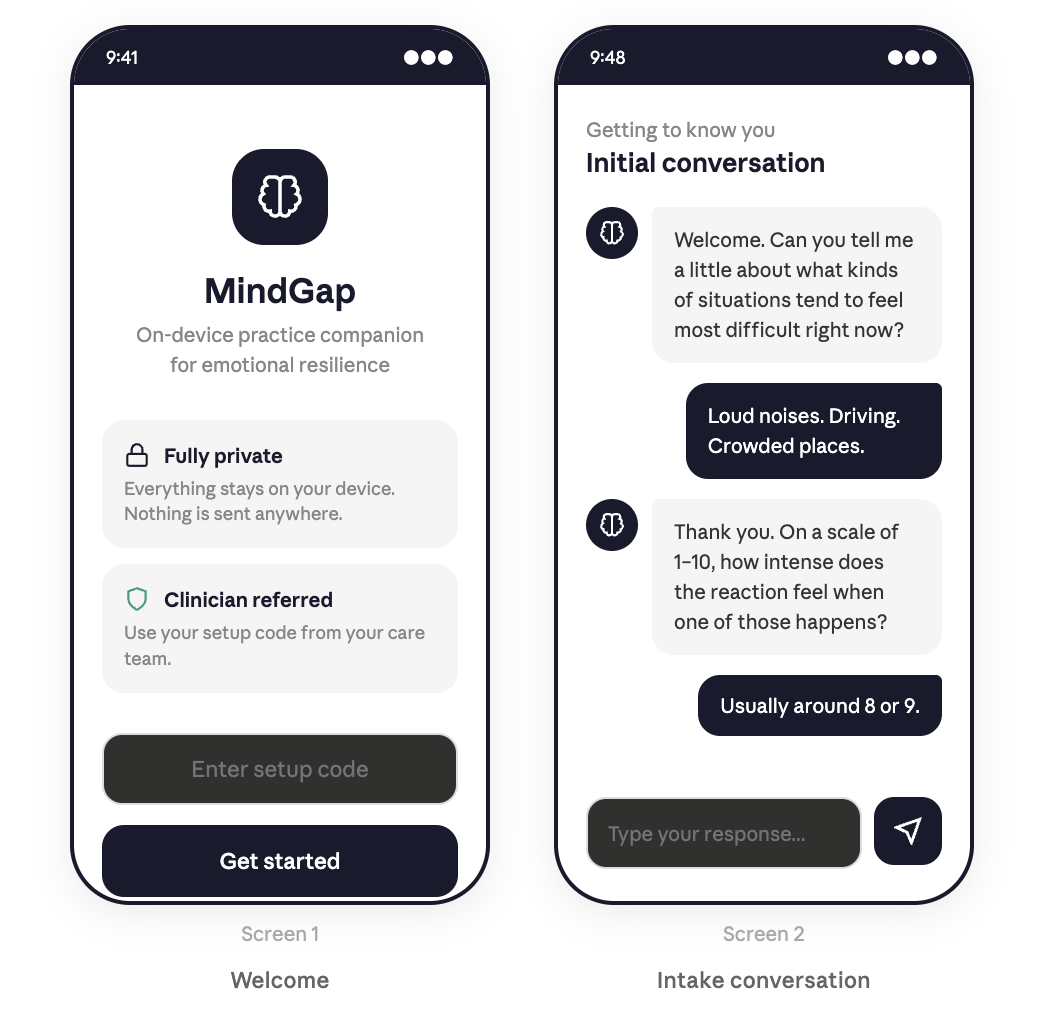}
\caption{MindGap onboarding and intake (Screens 1--2). The patient enters through a clinician-issued setup code on a clean, privacy-foregrounding welcome screen (Screen 1). The intake is conducted as a natural conversational dialogue — the agent asks open questions about triggers, symptom severity, and avoidance patterns, and the patient's responses drive the depth of the conversation (Screen 2). Intake data is stored locally and used to construct the personalised stimulus calibration ladder.}
\label{fig:screens:onboarding}
\end{figure}

Following orientation, the agent conducts a structured conversational intake — a natural dialogue rather than a clinical questionnaire, designed to minimise the administrative burden on a patient population for whom formal assessment processes may themselves be activating. The intake explores four domains: the nature of the traumatic experience at a level of detail the patient is comfortable providing; the primary triggers currently producing the strongest reactive responses; the current symptom severity and avoidance patterns; and the patient's prior experience with therapy and contemplative practice, which informs the initial ladder placement and the psychoeducation approach. Figure~\ref{fig:screens:onboarding} illustrates the welcome screen and a representative segment of the intake conversation — the agent's questions are open and gentle, the patient's responses drive the depth and direction of the dialogue, and the agent never pushes beyond what the patient offers.

Following intake, the agent constructs the patient's personalised stimulus calibration ladder locally on the device and delivers a brief psychoeducation module — explaining in plain language what feeling tone is, what the three layers of seeing involve, and what the patient can expect in sessions. No Buddhist terminology is used unless the patient requests it; the framework is presented entirely in neuroscience and plain English terms. The patient sets their session time preferences, notification style, and initial session length, completing the onboarding process.

\subsection{Daily Practice Sessions}
\label{sec:app:daily}

The daily practice session is the primary neuroplastic training environment of MindGap — the high-frequency repetition context in which the feeling tone gap intervention accumulates its therapeutic effect. Sessions run for five to ten minutes and are designed to be completable in ordinary daily contexts — before work, during a lunch break, or in the evening — without requiring a dedicated therapeutic space or extended preparation.

Each session begins with a brief check-in. The patient reports their current activation level on a simple scale and selects from a set of body sensation descriptors — tightness, tension, restlessness, calm — that orient their interoceptive attention before the first contact event is presented. This check-in serves a dual purpose: it establishes the patient's baseline state for that session, which the protocol engine uses to determine the appropriate stimulus level, and it begins the practice of present-moment interoceptive awareness that the subsequent feeling tone observation will build on.

Following a brief settling period — one or two minutes of guided breath awareness that primes the prefrontal regulatory circuit before amygdala engagement begins — the agent presents the first contact event at the patient's current ladder level. The presentation is brief, second-person, and present-tense: \textit{a quiet military base at sunrise, buildings, open ground, early light}. A deliberate pause follows before the feeling tone prompt arrives — allowing the affective signal to arise before any evaluative engagement is introduced. The agent then prompts: \textit{what do you notice right now? Is there a feeling of pleasant, unpleasant, or neutral?}

The patient selects or types their response, and the agent guides them through the three layers of seeing at a pace calibrated to the patient's engagement. At Layer 1, the agent confirms the affective label and invites further precision — \textit{you noticed tightness, can you stay with that for a moment}. At Layer 2, the agent introduces the decentering reframe — \textit{that feeling is arising in you right now, the scenario is just words, the feeling belongs to you, not to it}. At Layer 3, for patients who have demonstrated stable Layer 2 observation, the agent opens the implicit belief inquiry — \textit{what does this feeling seem to believe about this situation}. Each layer transition is made only when the patient's response indicates readiness, and the agent returns to a lower layer without comment if the patient's engagement indicates they are not yet ready to go deeper. Figure~\ref{fig:screens:practice} illustrates the daily check-in, the practice session interface with its layer progress indicator, and the real-world support screen.

Sessions close with a brief acknowledgement — the agent names what the patient observed and notes the practice that occurred — and the session data is logged locally. The patient is not given a performance evaluation. The only feedback is the observation that the practice happened, which is sufficient.

\subsection{Real-World Support and Weekly Deep Sessions}
\label{sec:app:realworld}

Two complementary session types extend the daily practice: real-world support for unscheduled activations in daily life, and weekly deep sessions for sustained higher-intensity therapeutic work.

\textbf{Real-world support} activates when the patient opens the application outside a scheduled session — typically at a moment of unscheduled activation in their daily environment. A veteran encountering a sudden loud noise at a supermarket, a trauma survivor in a crowded space, a first responder confronted with a scene reminiscent of a past incident — each of these is a moment when the reactive pathway fires in the environment where it was encoded. Real-world support is neuroplastically significant precisely because of this ecological validity: intercepting the reactive cascade in the real environment where the conditioned formation is active is more direct than practising interception in a structured session context \cite{foa2007prolonged}.

\begin{figure}[H]
\centering
\includegraphics[width=\textwidth]{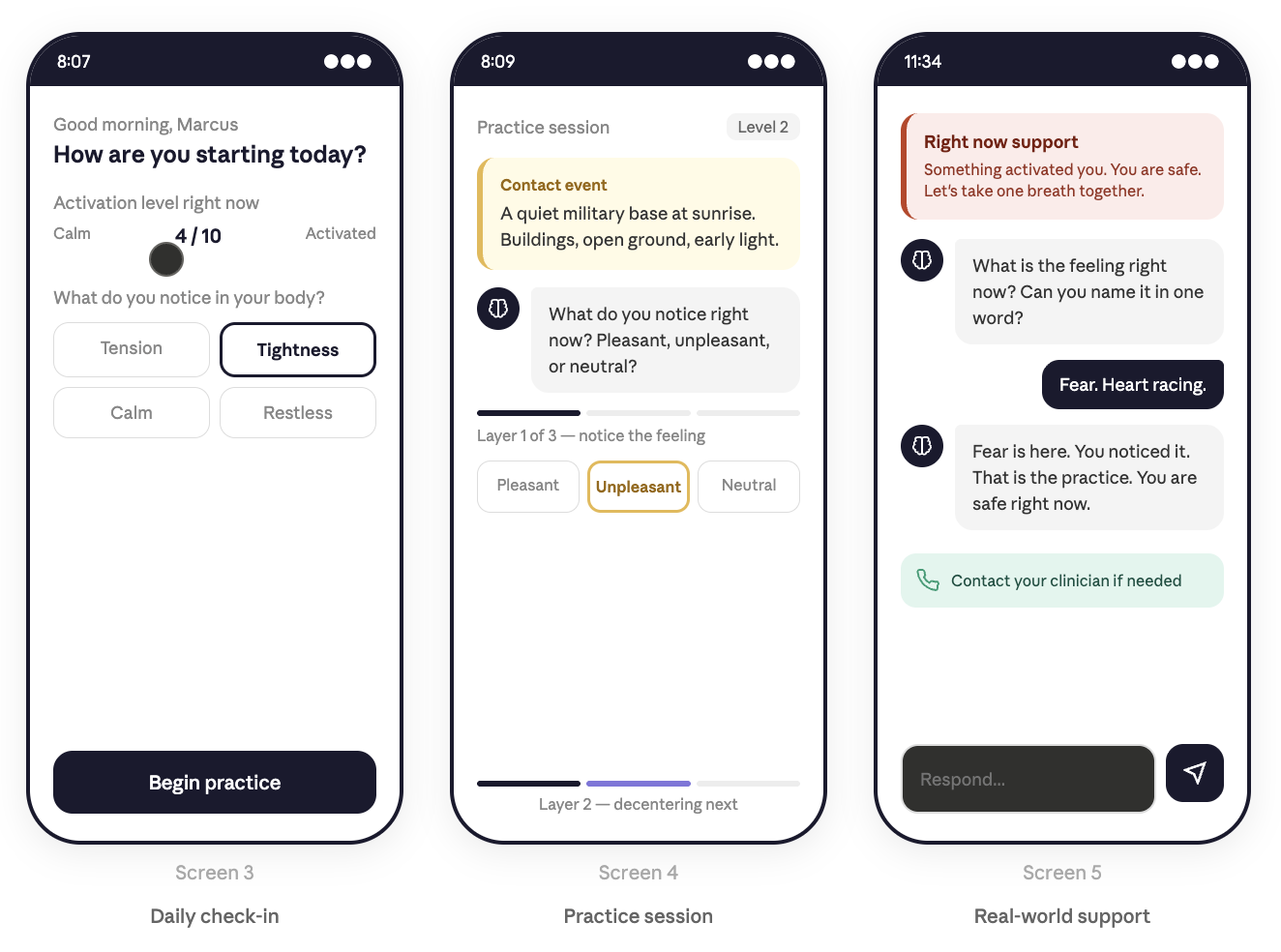}
\caption{MindGap daily practice journey (Screens 3--5). The daily check-in establishes activation level and body sensation before the first contact event (Screen 3). The practice session presents the contact event, the agent's feeling tone prompt, and a three-layer progress indicator — the patient selects their feeling tone response and the agent guides progressively deeper observation (Screen 4). Real-world support activates during unscheduled real-life activations — the interface is minimal, grounding-first, with a visible pathway to clinical contact (Screen 5).}
\label{fig:screens:practice}
\end{figure}

The real-world support interface is deliberately minimal — designed for use in a state of activation, when cognitive load must be kept low. The agent's opening acknowledges the activation directly: \textit{something activated you, you are safe, let's take one breath together}. From there the flow mirrors the daily session structure but compressed — one breath, one feeling tone observation, one layer of seeing if the patient is within the window of tolerance. The clinician contact option is visible throughout the real-world support flow, ensuring the patient always has a clear pathway to professional support if the activation exceeds what the app can safely hold. Figure~\ref{fig:screens:practice} (Screen 5) illustrates the real-world support interface.

\begin{figure}[H]
\centering
\includegraphics[width=0.85\textwidth]{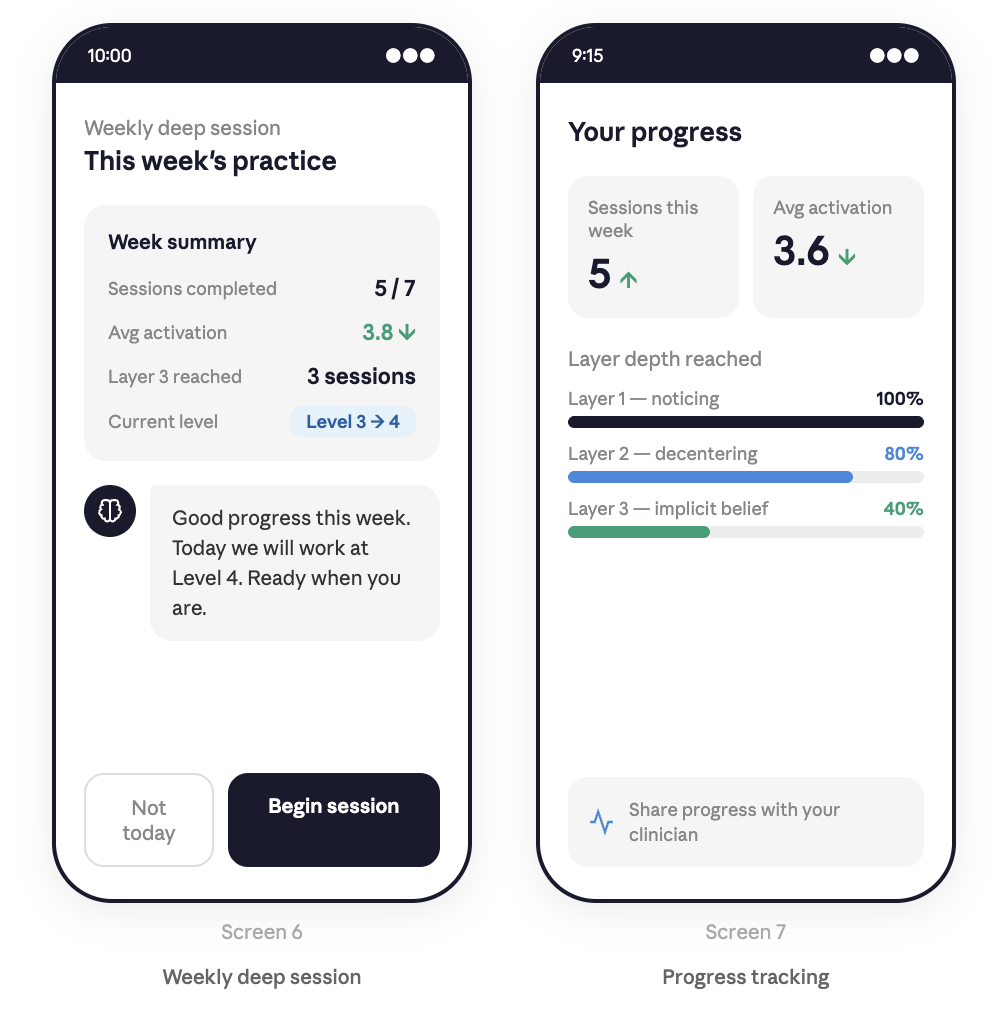}
\caption{MindGap progress review and weekly deep sessions (Screens 6--7). The weekly deep session opens with a structured summary of the week's practice — sessions completed, activation trajectory, layer depth reached, and ladder advancement status — before the agent briefs the patient on the session's intensity level (Screen 6). The progress screen displays the three neuroplastic proxies — activation trajectory, layer depth progression across all three layers, and an option to share the monthly summary with the clinical team through a patient-initiated export (Screen 7).}
\label{fig:screens:progress}
\end{figure}

\textbf{Weekly deep sessions} run for thirty to forty-five minutes and serve three functions within the therapeutic arc. First, they provide a structured weekly review of the patient's practice trajectory — the agent opens each weekly session by summarising the week's sessions, activation trajectory, layer depth reached, and ladder position, giving the patient a regular view of their neuroplastic progress. Second, they advance the therapeutic intensity — weekly sessions operate at a higher stimulus level than daily practice, introducing the patient to ladder levels that the daily sessions have prepared them for but not yet accessed. Third, they provide a space for deeper Layer 3 exploration — the implicit belief inquiry that daily sessions may touch but rarely sustain. Figure~\ref{fig:screens:progress} (Screen 6) illustrates the weekly session opening with its week summary and agent briefing before the session begins.

\subsection{Progress Review and Clinical Integration}
\label{sec:app:progress}

Longitudinal progress tracking is central to MindGap's therapeutic design — not as performance measurement but as neuroplastic visibility. The progress screen gives patients a concrete view of the changes the practice is producing: activation levels trending downward, layer depth access increasing, stimulus levels advancing. This visibility serves a therapeutic function beyond data tracking — it makes the neuroplastic work tangible, providing evidence of change that sustains motivation through the inevitable periods of plateau and regression that sustained practice involves.

Three neuroplastic proxies are displayed prominently. Activation trajectory shows the trend in session-opening activation levels over time — a declining trend is consistent with LTD-mediated weakening of the reactive pathway and the consequent rise in amygdala threshold. Layer depth progression shows the proportion of sessions in which each layer was reached — increasing Layer 2 and Layer 3 access reflects the LTP-mediated strengthening of the prefrontal observing pathway. A third metric, available in the detailed view, shows the stimulus level currently being worked at — advancement up the ladder reflects the patient's growing capacity to observe at the feeling tone gap under progressively more activating conditions.

Monthly the application generates a structured progress summary — a concise report covering session frequency, activation trajectory, layer depth progression, stimulus level reached, and the stimulus categories producing the highest and lowest activation. This summary is presented to the patient within the application and can be exported through a patient-initiated sharing mechanism — the patient selects what to share and with whom, and the data leaves the device only through this explicit action. The clinical team uses the summary to inform ladder progression decisions, identify stimulus categories requiring focused attention in formal therapy, assess whether pharmacotherapy adjustment or therapy intensity modification may be indicated, and track the patient's neuroplastic trajectory relative to the trial's outcome measures.

Step-down from active practice to maintenance mode occurs when three criteria are met across consecutive months: activation levels at equivalent stimulus intensities have declined to within the normal range, Layer 3 access has become consistent across sessions, and the patient reports sustained improvement in daily functioning. In maintenance mode, session frequency reduces from daily to three times per week, real-world practice becomes the primary mode, and monthly check-ins with the clinical team use the app's progress data to monitor stability and respond to regression when it occurs.

\section{Evaluation: A Clinical Use Case Analysis}
\label{sec:evaluation}

MindGap is a proposed framework whose neuroplastic claims require empirical validation through controlled longitudinal study. The present section does not substitute for that validation. Rather, it traces the complete therapeutic arc of a representative clinical use case — a combat veteran with moderate-severe PTSD — to demonstrate the framework's clinical coherence and show, in concrete detail, how the dependent origination chain unfolds and is progressively interrupted across the course of MindGap practice. The use case is illustrative rather than empirical; the interaction dialogues, activation trajectories, and neuroplastic proxy values presented below are representative of the framework's intended operation and grounded in the theoretical mechanisms established in Section~\ref{sec:framework}, but do not constitute observed clinical data. Section~\ref{sec:evaluation:limitations} addresses the limitations of this approach and outlines the empirical validation pathway.

\subsection{Clinical Profile}
\label{sec:evaluation:profile}

Marcus is a 34-year-old combat veteran who served two tours in Afghanistan. During his second tour, he was the sole survivor of a roadside IED explosion that killed two members of his unit. He was diagnosed with combat-related PTSD six months following his return, with a PCL-5 score of 58 indicating moderate-severe symptom severity. His primary triggers are sudden loud sounds, vehicles on roads, and crowded spaces — all stimuli associated with the environments in which the traumatic event and its immediate aftermath occurred. Current symptoms include hyperarousal with sleep disturbance, active avoidance of driving and crowded environments, and intrusive re-experiencing through both flashbacks and nightmares. He has been engaged in weekly therapy sessions at Blanchfield Army Community Hospital for four months with partial symptom reduction, and is referred to MindGap as a between-session neuroplastic practice tool to complement his ongoing clinical care.

From the dependent origination perspective, Marcus's trauma has produced extreme gathi — trauma-encoded implicit beliefs, encoded through Hebbian long-term potentiation at maximum intensity during the traumatic event, that certain stimuli mean mortal danger. These beliefs operate below conscious awareness as structural features of his amygdala's threat-evaluation system, lowering the activation threshold so that stimuli merely associated with the original trauma now trigger the full HPA stress cascade before conscious appraisal has engaged \cite{pitman2012biological, mahan2012synaptic}. The dependent origination chain fires rapidly and completely at each triggering event: contact, extreme unpleasant feeling tone, full aversion mobilisation, re-experiencing narrative, and reactive behavior — each cycle deepening the conditioned formation through long-term potentiation. The MindGap intervention targets this chain at the feeling tone gap, aiming to intercept the cascade before aversion fully mobilises and to progressively dissolve the trauma-encoded implicit beliefs at Layer 3.

\subsection{Intake and Stimulus Ladder Construction}
\label{sec:evaluation:intake}

During intake, the agent conducts a structured conversational dialogue to map Marcus's trigger profile, current symptom severity, and avoidance patterns. Marcus describes his primary triggers — loud noises, driving, crowded places — and rates their current activation intensity at approximately 8 to 9 out of 10. He reports active avoidance of supermarkets, public transport, and all driving. He has no prior contemplative practice experience. The agent frames the practice in plain neurobiological terms — explaining that the sessions will involve noticing an immediate feeling before it becomes a reaction, and that this noticing, repeated across many sessions, changes how the brain responds over time. No Buddhist terminology is introduced.

From Marcus's intake profile, the protocol engine constructs his personalised stimulus calibration ladder:

\begin{itemize}
\item \textbf{Level 1 — Conceptual:} ``Think about the word deployment for a moment.''
\item \textbf{Level 2 — Peripheral:} ``A quiet military base at sunrise. Buildings, open ground, early light.''
\item \textbf{Level 3 — Adjacent:} ``A soldier is cleaning equipment after a long patrol.''
\item \textbf{Level 4 — Direct, mild:} ``You hear distant vehicles approaching on a road ahead.''
\item \textbf{Level 5 — Moderately activating:} ``There is a sudden loud bang outside your building.''
\item \textbf{Level 6 — Strongly activating:} ``You are in a vehicle approaching an uncertain road ahead.''
\end{itemize}

The protocol engine places Marcus at Level 1 for his first two weeks of daily practice, with advancement to Level 2 contingent on demonstrating stable Layer 1 observation across three consecutive sessions at Level 1.

\subsection{Early Sessions — Weeks One to Four}
\label{sec:evaluation:early}

\begin{figure}[H]
\centering
\includegraphics[width=\textwidth]{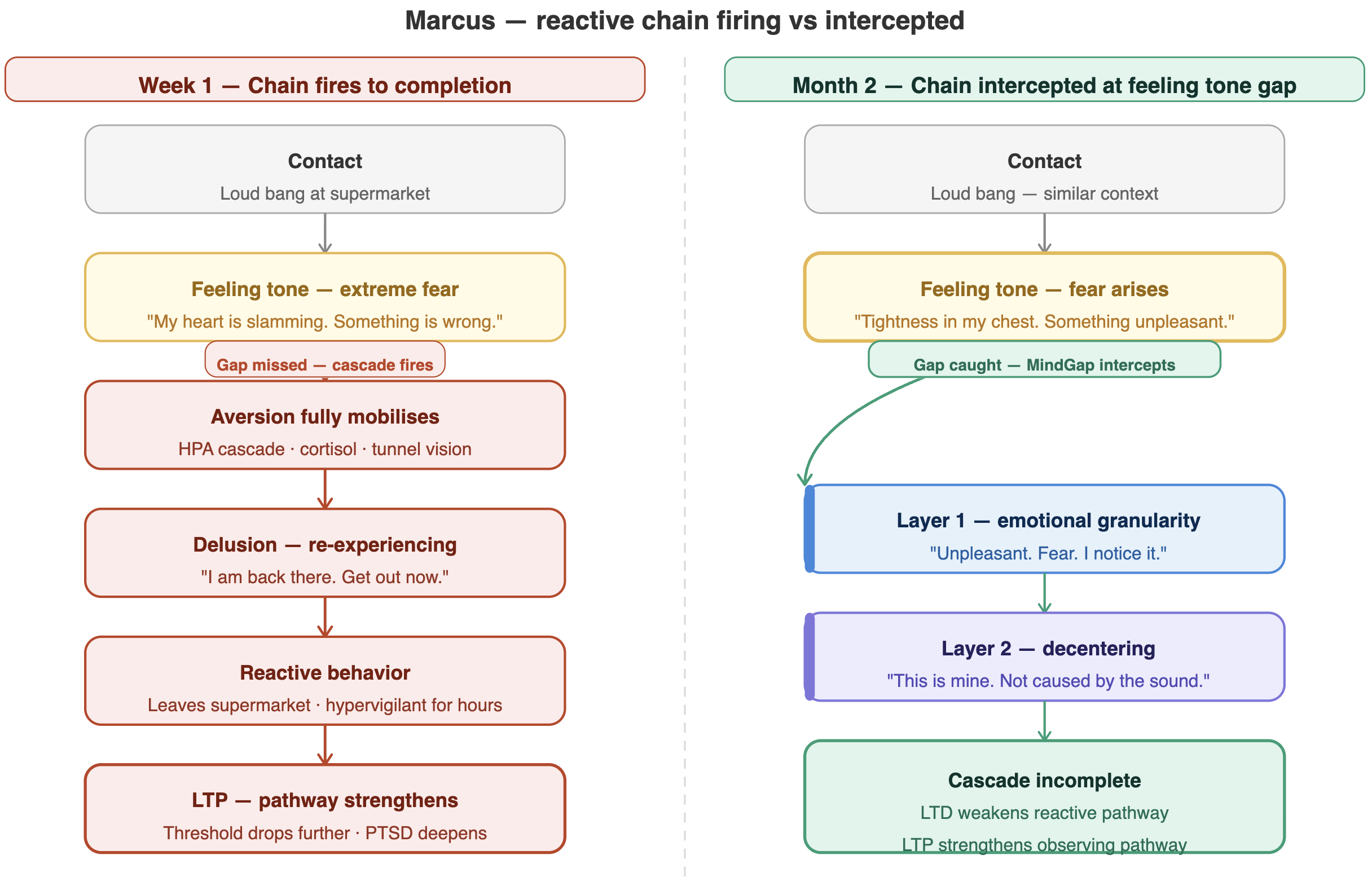}
\caption{The dependent origination chain in Marcus's early sessions versus month two. Left: in week one the chain fires to completion — contact through feeling tone, aversion, re-experiencing delusion, and reactive behavior — strengthening the PTSD pathway through long-term potentiation. Right: by month two Marcus catches the feeling tone gap before aversion fully mobilises, accessing Layer 1 emotional granularity and Layer 2 decentering. The reactive pathway does not complete its activation, initiating LTD-mediated weakening while the observing pathway strengthens through LTP.}
\label{fig:marcus:chain}
\end{figure}

Marcus's early sessions are characterised by the pattern most common in the first weeks of MindGap practice: the dependent origination chain fires to completion before awareness has established itself at the feeling tone gap, and noticing arrives retrospectively — after the reaction has already mobilised — rather than upstream. Figure~\ref{fig:marcus:chain} illustrates this pattern alongside its month-two contrast.

In a representative week-two session, the agent presents a Level 1 contact event — the word ``deployment'' — and prompts feeling tone observation. Before Marcus can form a response, he reports that he is already in the reaction: thinking about his unit, feeling the familiar tightening, his mind already constructing the narrative of the day of the explosion. The feeling tone has arisen and been immediately followed by aversion and the beginning of re-experiencing delusion — the chain has fired without the gap becoming visible. The agent's response is calibrated precisely to this pattern: it validates the retrospective noticing as the beginning of the practice rather than a failure, and names what happened in the chain terms the patient can use — ``you noticed you were already in the reaction; that noticing, even after the fact, is the observing pathway activating.'' This framing is therapeutically important: retrospective recognition is neuroplastically real, and treating it as failure would deepen aversion toward the practice itself rather than dissolving the reactive tendency it targets.

By week three, Marcus begins reporting that he can sometimes notice the tightening before it has fully elaborated into the re-experiencing narrative. The gap is not yet accessible before aversion mobilises, but the interval between feeling tone and full narrative elaboration is becoming perceptible. Session-opening activation levels, which began at approximately 6.8 in week one, show early decline to around 6.2 by week four — consistent with the beginning of LTD-mediated weakening of the reactive pathway at the lower stimulus intensities the early sessions employ.

\subsection{Developing Practice — Months One to Three}
\label{sec:evaluation:developing}

The transition from month one to month two marks the first consistent appearance of the feeling tone gap as a workable moment in Marcus's practice. The gap does not widen dramatically — it becomes perceptible rather than imperceptible, available rather than inaccessible. Marcus begins catching the unpleasant feeling tone at Level 2 contact events before it has propagated fully into the aversion cascade, noticing the tightness in his chest as a discrete event — ``something unpleasant is here'' — rather than as the beginning of a reactive sequence he is already inside.

\begin{figure}[H]
\centering
\includegraphics[width=\textwidth]{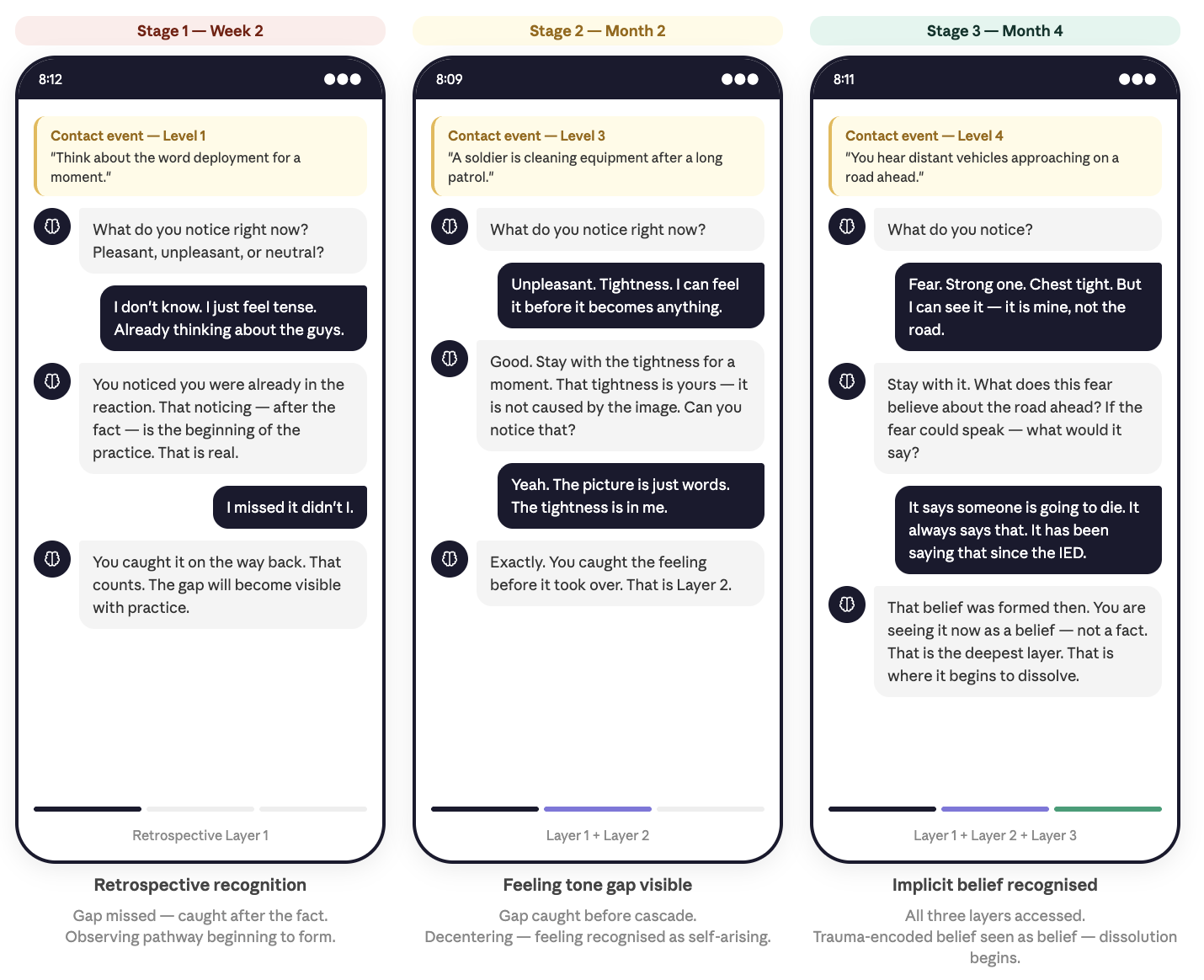}
\caption{Agent-patient dialogue across three stages of MindGap practice. Stage 1 (Week 2): Marcus misses the feeling tone gap and reacts before noticing — the agent validates retrospective recognition as the beginning of practice rather than a failure. Stage 2 (Month 2): Marcus catches the feeling tone before the cascade completes and reaches Layer 2 decentering — recognising the tightness as self-arising rather than caused by the stimulus. Stage 3 (Month 4): All three layers are accessed — Marcus identifies the trauma-encoded implicit belief beneath the feeling for the first time, naming it as the belief formed during the IED explosion. The deepest therapeutic target of the MindGap protocol becomes visible and workable.}
\label{fig:marcus:dialogue}
\end{figure}

In a representative session from month two, the agent presents a Level 3 contact event — a soldier cleaning equipment after a patrol — and prompts feeling tone observation. Marcus responds that he notices unpleasantness, tightness, ``I can feel it before it becomes anything.'' The agent confirms the Layer 1 observation and guides the Layer 2 transition — ``that tightness is yours, it is not caused by the image, can you notice that?'' Marcus's response — ``the picture is just words, the tightness is in me'' — constitutes the decentering recognition that Layer 2 targets: the feeling is self-arising, not a property of the stimulus. This is a neuroplastically significant moment. The cascade has been intercepted before full HPA activation. The reactive pathway has not fired to completion. The LTD mechanism begins to operate on the synaptic connections that would otherwise have propagated the cascade from aversion to full stress response. Figure~\ref{fig:marcus:dialogue} presents this dialogue alongside the week-two and month-four exchanges, showing the progressive development of the practice across the three stages.

The supermarket incident occurs in week seven. Marcus is in a crowded supermarket when a shopping cart crashes loudly nearby. The familiar cascade fires immediately — heart rate surges, tunnel vision begins, the impulse to leave activates. But he opens MindGap. The real-world support interface presents the immediate grounding prompt, and Marcus names the feeling: ``fear, heart racing.'' The agent confirms the naming — ``fear is here, you noticed it, that is Layer 1, you are safe right now'' — and Marcus stays in the supermarket. He does not complete the full reactive behavior of leaving. The cascade has been partially intercepted in a real-world activation context — the most ecologically significant neuroplastic event of his practice to this point, occurring in the environment where the conditioned formation is most strongly encoded \cite{foa2007prolonged}. By the end of month two, session-opening activation levels have declined to approximately 5.0, Layer 2 is accessible in around sixty percent of sessions, and the protocol engine advances Marcus to Level 3 for his daily practice.

\subsection{Deepening Practice — Months Three to Six}
\label{sec:evaluation:deepening}

The third and fourth months bring the most significant qualitative shift in Marcus's practice: the first consistent access to Layer 3 — the metacognitive recognition of the trauma-encoded implicit belief beneath the feeling tone. This shift does not arrive suddenly. It emerges gradually from the consolidation of Layer 2 observation — once the feeling is reliably seen as self-arising rather than stimulus-caused, the question of what conditioned belief is generating it becomes accessible rather than opaque.

In a weekly deep session in month four, the agent presents a Level 4 contact event — distant vehicles approaching on a road ahead — and Marcus catches the feeling tone immediately: ``fear, strong one, chest tight, but I can see it, it is mine, not the road.'' The agent moves to the Layer 3 prompt — ``what does this fear believe about the road ahead, if the fear could speak, what would it say?'' — and Marcus's response marks the therapeutic inflection point of his case: ``it says someone is going to die, it always says that, it has been saying that since the IED.'' This is the implicit belief — the trauma-encoded conditioned formation that has been generating the extreme feeling tone at every triggering contact event since the traumatic encoding. Marcus is not articulating a conscious cognition. He is recognising, for the first time with some clarity, the belief that has been operating below awareness in his amygdala's threat-evaluation system since the day of the explosion. The agent's response names what has happened without amplifying it — ``that belief was formed then, you are seeing it now as a belief, not a fact, that is where it begins to dissolve'' — and the session closes. Marcus reports feeling ``lighter, tired, but something shifted.''

This Layer 3 recognition is the deepest neuroplastic event MindGap targets. The implicit belief — previously operating as an automatic structural feature of the amygdala's threat-evaluation system — has been observed as a conditioned formation rather than an accurate perception of current reality. Teasdale's research identifies this metacognitive awareness of implicit belief as the layer of practice most strongly associated with durable reduction in reactive relapse, because it addresses the conditioned formation that generates the reactive cascade rather than merely interrupting its immediate expression \cite{teasdale2002metacognitive}. In neuroplastic terms, the LTD mechanism now operates not only on the reactive cascade links downstream of feeling tone, but on the implicit belief substrate from which the extreme feeling tone itself arises — the deepest structural target of the MindGap intervention. By the end of month four, session-opening activation levels have declined to approximately 3.8, Layer 3 is accessible in approximately forty percent of sessions, and Marcus advances to Level 4 stimuli in his weekly deep sessions.

\subsection{Evaluation Against Clinical and Neuroplastic Criteria}
\label{sec:evaluation:criteria}

Evaluated against five criteria grounded in the theoretical framework of Section~\ref{sec:framework}, the MindGap application to Marcus's case demonstrates the following:

\textbf{Upstream intervention.} The protocol successfully delivers contact events and prompts feeling tone observation before the cascade completes across the majority of sessions by month two. The real-world supermarket incident demonstrates the real-world transferability of the upstream interception — the most demanding test of the practice's ecological validity.

\textbf{Three-layer therapeutic depth.} All three layers are accessed progressively across the twelve-week trajectory. Layer 1 is established within the first two weeks. Layer 2 becomes consistent from month two. Layer 3 first appears in month four and consolidates progressively thereafter — consistent with the theoretical prediction that each layer requires the preceding layer to be stabilised before it becomes accessible.

\textbf{Window of tolerance adherence.} The stimulus ladder and step-back protocol maintain Marcus within the therapeutic window of tolerance across all sessions. No sessions trigger acute dissociation or crisis-level activation. The real-world incident in week seven approaches the upper boundary of the window but does not breach it — the immediate real-world support interface provides sufficient grounding.

\textbf{Neuroplastic proxy trajectory.} Session-opening activation levels decline from 6.8 in week one to 3.8 by week twelve — a reduction of 44\% at equivalent or higher stimulus intensity, consistent with the predicted LTD-mediated weakening of the reactive pathway and the consequent rise in amygdala threshold. Layer depth progression follows the theoretical prediction: Layer 1 consistent from week one, Layer 2 accessible from week four, Layer 3 first appearing in week nine. Stimulus level advances from Level 1 through Level 4 across the twelve weeks, reflecting growing observing capacity under progressively more activating conditions.

\textbf{Clinical integration.} The monthly progress summary shared with Marcus's clinical team informs three specific therapy decisions across the twelve weeks: advancement of the in-session exposure to Level 4 vehicle scenarios in week ten; identification of crowded spaces as the stimulus category requiring additional clinical attention based on app log data; and a discussion of whether continued pharmacotherapy is indicated given the neuroplastic proxy trajectory. The app and the clinical care pathway operate as complements — the app providing the high-frequency between-session neuroplastic training that weekly therapy sessions cannot deliver, and the clinical team providing the oversight, interpretive context, and higher-intensity processing that the app is not designed to hold.

\subsection{Limitations and Empirical Validation Pathway}
\label{sec:evaluation:limitations}

The use case analysis presented in this section is illustrative rather than empirical. Marcus is a representative clinical profile constructed to demonstrate the framework's operation across a realistic therapeutic arc — not a trial participant whose outcomes have been measured. The activation trajectory values, layer depth progression data, and dialogue exchanges are theoretically grounded representations of the framework's intended behaviour rather than observed clinical data. The use case cannot substitute for controlled empirical study and should not be interpreted as evidence of clinical efficacy.

Individual variation in trauma profile, symptom severity, prior contemplative experience, neurobiological baseline, and therapeutic context will produce significant variation from the trajectory described here. Patients with more severe PTSD, complex trauma histories, or significant comorbidities may progress more slowly, require more intensive clinical oversight, or find the window of tolerance more difficult to maintain. The stimulus ladder and protocol engine are designed to accommodate this variation, but the calibration logic itself requires empirical validation to establish optimal advancement criteria and step-back thresholds for different patient populations.

Empirical validation of MindGap's neuroplastic claims requires a randomised controlled trial comparing MindGap plus standard care against standard care alone in a military or veteran PTSD population. Primary neuroplastic outcome measures should include amygdala reactivity assessed through fMRI, prefrontal-amygdala functional connectivity, and heart rate variability as a physiological proxy for autonomic regulatory capacity. Primary symptomatic outcome measures should include the PCL-5 and the clinician-administered CAPS-5. Behavioural biomarkers from the app's session logs — activation trajectory, layer depth progression, and feeling tone recognition latency — should be included as secondary outcomes, providing a novel computational measure of neuroplastic progress that does not require neuroimaging. The clinical sites of Blanchfield Army Community Hospital and McDonald Army Health Center represent established DoD clinical research contexts where such a trial could be conducted with appropriate ethical oversight and access to the target patient population.

\section{Conclusion and Future Work}
\label{sec:conclusion}

Post-Traumatic Stress Disorder is not a psychological response that time heals — it is a neuroplastic encoding that time deepens. The clinical approaches that currently constitute the standard of care — prolonged exposure, EMDR, cognitive behavioural therapy — produce genuine improvements, but all operate downstream of the reactive cascade, managing its expression rather than dissolving the pathway that generates it. MindGap proposes a different therapeutic logic: upstream dissolution at the feeling tone gap, guided by the dependent origination framework and delivered through a fine-tuned on-device conversational AI agent that guides patients through three progressive layers of observation — emotional granularity, decentering, and metacognitive recognition of the trauma-encoded implicit belief beneath the fear. Running entirely on the patient's device with no data egress, it is deployable in the military and sensitive clinical contexts where existing cloud-based solutions are not permitted.

The clinical use case analysis demonstrates the framework's coherence across a realistic therapeutic arc — showing how the dependent origination chain is progressively intercepted, how the three layers emerge across months of practice, and how the trauma-encoded implicit belief becomes visible at Layer 3 in a way that existing approaches do not reach. These remain illustrative demonstrations. Controlled longitudinal validation — a randomised controlled trial measuring amygdala reactivity, prefrontal-amygdala connectivity, PCL-5 and CAPS-5 symptom severity, and behavioural biomarkers from the app's session logs — is the immediate priority before MindGap can be recommended as a clinical intervention.

Future directions include physiological sensor integration for objective window-of-tolerance monitoring, extension to other trauma populations and adjacent clinical conditions including anxiety and substance use disorder, virtual reality integration for higher-intensity exposure, and application of the upstream dissolution framework to high-performance non-clinical contexts including elite athletic performance and military decision-making under pressure. The dependent origination framework that underlies MindGap has been refined across 2,500 years of contemplative investigation. What is new is the convergence of neuroplasticity science, conversational AI, and on-device inference that makes its clinical operationalisation possible — and necessary — for the first time.



\bibliographystyle{elsarticle-num}
\bibliography{reference}

\end{document}